\pdfoutput=1

\documentclass[11pt]{article}

\usepackage{naacl2021}

\usepackage{times}
\usepackage{latexsym}

\usepackage[T1]{fontenc}

\usepackage[utf8]{inputenc}

\usepackage{microtype}
\usepackage{times,latexsym}
\usepackage{url}
\usepackage{times}
\usepackage{latexsym}
\usepackage{array}
\usepackage{tikz}
\usepackage{soul}
\usepackage{caption, floatrow}

\usepackage{times}
\usepackage{latexsym}
\usepackage{graphicx}
\usepackage{multirow}
\usepackage{amssymb}  
\usepackage{amsthm}
\usepackage{amsmath}
\usepackage[official]{eurosym}
\usepackage{mathtools}

\usepackage{arydshln}
\usepackage{algorithm}
\usepackage{algorithmic}
\usepackage{graphicx}
\usepackage{booktabs}
\usepackage{calrsfs}
\usepackage{array}
\usepackage[T1]{fontenc}
\usepackage{CJKutf8}
\usepackage[OT2, T1]{fontenc}
\usepackage[russian, english]{babel}
\usepackage[T1]{fontenc}
\usepackage{fdsymbol}
\usepackage{paralist}

\usepackage[utf8]{inputenc}

\title{Instructions for NAACL-HLT 2021 Proceedings}

\usepackage{xspace,mfirstuc,tabulary}

\definecolor{bblue}{HTML}{4F81BD}
\definecolor{oorange}{HTML}{F4C842}
\definecolor{rred}{HTML}{C0504D}
\definecolor{ggreen}{HTML}{9BBB59}
\definecolor{ppurple}{HTML}{9F4C7C}
\definecolor{darkgreen}{HTML}{228B22}
\definecolor{cred}{HTML}{D81B60}
\definecolor{cblue}{HTML}{1E88E5}
\definecolor{cyellow}{HTML}{FFC107}
\definecolor{nred}{HTML}{e41a1c}
\definecolor{nblue}{HTML}{377eb8}
\definecolor{ngreen}{HTML}{4daf4a}
\definecolor{lblue}{HTML}{6C8EBF}

\usepackage{xspace}
\newcommand{\tydi}{\textsc{TyDi~QA}\xspace}
\newcommand{\xor}{\textsc{Xor} QA\xspace}
\newcommand{\tydixor}{\textsc{Xor}-\textsc{TyDi} QA\xspace}
\newcommand{\xorretrieve}{\textsc{XOR-Retrieve}\xspace}
\newcommand{\xorengspan}{\textsc{Xor-EnglishSpan}\xspace}
\newcommand{\xorfull}{\textsc{Xor-Full}\xspace}
\newcolumntype{L}[1]{>{\raggedright\let\newline\\\arraybackslash\hspace{0pt}}m{#1}}
\newcolumntype{C}[1]{>{\centering\let\newline\\\arraybackslash\hspace{0pt}}m{#1}}
\newcolumntype{R}[1]{>{\raggedleft\let\newline\\\arraybackslash\hspace{0pt}}m{#1}}
\usepackage{pifont}
\newcommand{\xmark}{\textcolor{red}{\ding{55}}}
\newcommand{\cmark}{\textcolor{darkgreen}{\ding{51}}}

\title{\xor: Cross-lingual Open-Retrieval Question Answering}

\author{
  \parbox{0.6\linewidth}{\centering Akari Asai$^{\clubsuit}$, Jungo Kasai$^{\clubsuit}$, Jonathan H. Clark$^{\vardiamondsuit}$, Kenton Lee$^{\vardiamondsuit}$, Eunsol Choi$^{\varheartsuit}$, Hannaneh Hajishirzi$^{\clubsuit\spadesuit}$} \\
 $^\clubsuit$University of Washington~~$^\vardiamondsuit$Google Research \\ 
 $^\varheartsuit$The University of Texas at Austin~~$^\spadesuit$Allen Institute for AI \\
  \texttt{\{akari, jkasai, hannaneh\}@cs.washington.edu}  \\
  \texttt{\{jhclark, kentonl\}@google.com},~~\texttt{eunsol@cs.utexas.edu} \\
}

\date{}

\begin{document}
\maketitle

\begin{abstract}
Multilingual question answering tasks typically assume that answers exist in the same language as the question. Yet in practice, many languages face both \textit{information scarcity}---where languages have few reference articles---and \textit{information asymmetry}---where questions reference concepts from other cultures.
This work extends open-retrieval question answering to a cross-lingual setting enabling questions from one language to be answered via answer content from another language.
We construct a large-scale dataset built on 40K information-seeking questions across 7 diverse non-English languages that \tydi could not find same-language answers for. 
Based on this dataset, we introduce a task framework, called \textbf{Cross}-lingual \textbf{O}pen-\textbf{R}etrieval \textbf{Q}uestion \textbf{A}nswering (\textbf{\xor}), that consists of three new tasks involving cross-lingual document retrieval from multilingual and English resources.
We establish baselines with state-of-the-art machine translation systems and cross-lingual pretrained models. Experimental results suggest that \xor~is a challenging task that will facilitate the development of novel techniques for multilingual question answering.
Our data and code are available at \url{https://nlp.cs.washington.edu/xorqa/}.
\end{abstract}

\section{Introduction}

Information-seeking questions---questions from people who are actually looking for an answer---have been increasingly studied in question answering (QA) research.
Fulfilling these information needs has led the research community to look further for answers: beyond paragraphs and articles toward performing \textbf{open retrieval}\footnote{We use  \textbf{open retrieval}---instead of \textbf{open domain}---to refer to models that can access answer context from large document collections. We avoid using open domain due to its double meaning as ``covering topics from many domains.''} on large-scale document collections~\cite{chen-yih-2020-open}. Yet the bulk of this work has been exclusively on English. In this paper, we bring together for the first time information-seeking questions, open-retrieval QA, and multilingual QA to create a multilingual open-retrieval QA dataset that enables \textit{cross}-lingual answer retrieval.
\begin{figure}[t]
  \includegraphics[width=\linewidth]{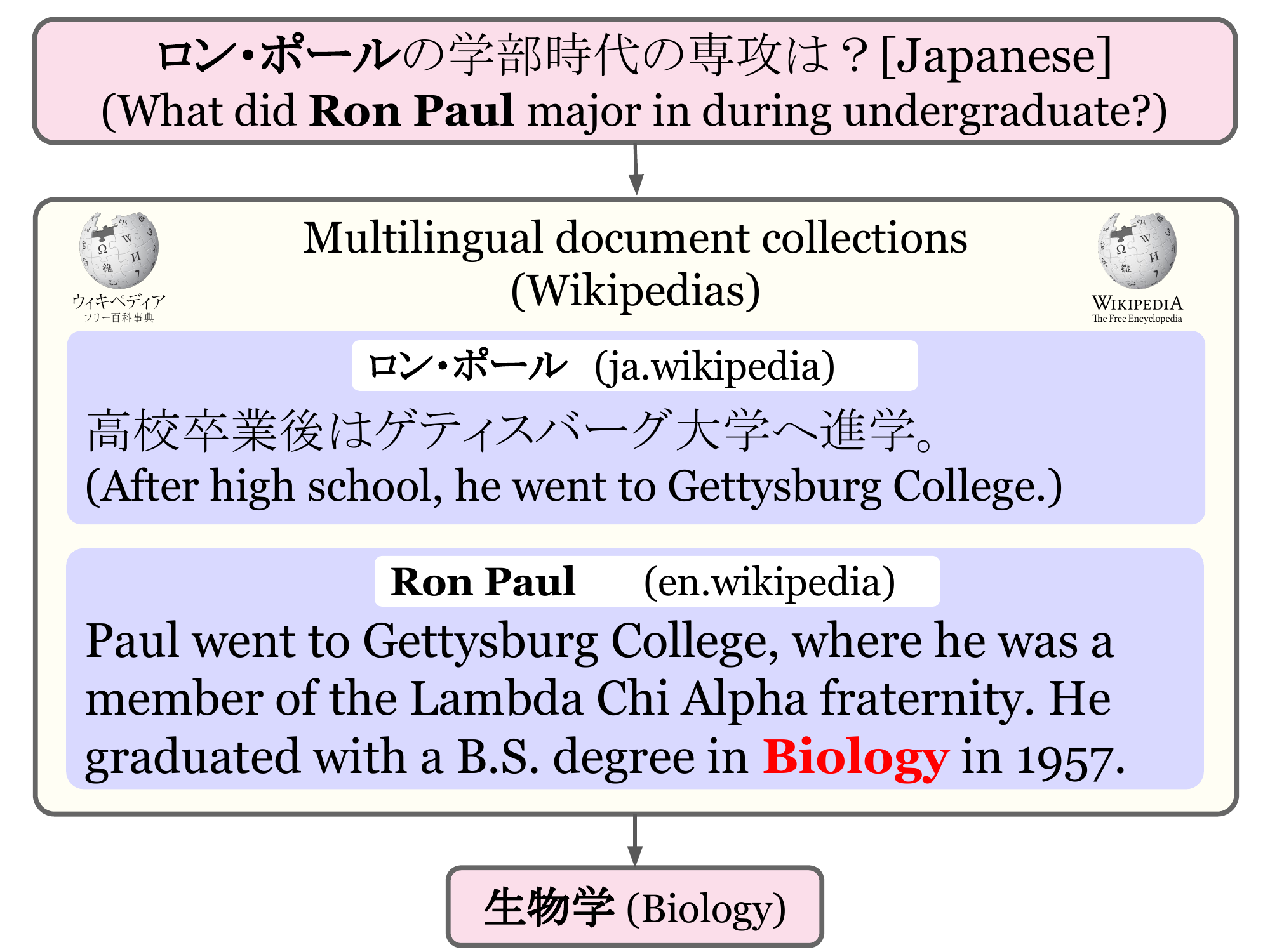}
  \caption{Overview of \xor. Given a question in $L_i$, the model finds an answer in either English or $L_i$ Wikipedia and returns an answer in English or $L_i$. $L_i$ is one of the 7 typologically diverse languages. 
  }
  \label{img:overview}
\end{figure}

While multilingual open QA systems would benefit the many speakers of non-English languages, there are several pitfalls in designing such a dataset. First, a multilingual QA dataset should include questions from non-English native speakers to represent real-world applications.
Questions in most recent multilingual QA datasets~\cite{lewis2019mlqa,Artetxe:etal:2019,mkqa} are translated from English, which leads to English-centric questions such as questions about American sports, cultures and politics. Second, it is important to support retrieving  answers in  languages other than the original language due to {\it information scarcity} of low-resource languages~\cite{internetstat}. Moreover, questions strongly related to entities from other cultures are less likely to have answer content in the questioner's language due to cultural bias~(\textit{information asymmetry}, \citealp{callahan2011cultural}).
For example, Fig.~\ref{img:overview} shows that the Japanese Wikipedia article of an American politician, Ron Paul, does not have information about his college degree perhaps because Japanese Wikipedia editors are less interested in specific educational backgrounds of American politicians.

In this paper, we introduce the task of cross-lingual open-retrieval question answering (\xor) which aims at answering multilingual questions from non-English native speakers given multilingual resources. 
To support research in this area, we construct a dataset (called \tydixor) of 40k annotated questions and answers across 7 typologically diverse languages. 
Questions in our dataset are inherited from \tydi~\cite{tydiqa}, which are written by native speakers and are originally unanswerable due to the information scarcity or asymmetry issues. 
\tydixor is the first large-scale cross-lingual open-retrieval QA dataset that consists of information-seeking questions from native speakers and multilingual reference documents.
 
\tydixor is constructed with an annotation pipeline that allows for cross-lingual retrieval from large-scale Wikipedia corpora (\S\ref{sec:dataset}). Unanswerable questions in \tydi are first translated into English by professional translators. 
Then, annotators find answers to translated queries given English Wikipedia using our new model-in-the-loop annotation framework that reduces annotation errors.
Finally, answers are verified and  translated  back to the target languages. 

Building on the dataset, we introduce three new tasks in the order of increasing complexity (\S\ref{sec:task}).
In \xorretrieve, a system retrieves English Wikipedia paragraphs with sufficient information to answer the question posed in the target language. 
\xorengspan takes one step further and finds a minimal answer span from the retrieved English paragraphs.
Finally, \xorfull expects a system to generate an answer end to end in the target language by consulting both English and the target language's Wikipedia.
\xorfull~is our ultimate goal, and the first two tasks enable researchers to diagnose  where their models fail and develop under less coding efforts and resources.

We provide baselines that extend state-of-the-art open-retrieval QA systems~\cite{Asai2020Learning,karpukhin2020dense} to our multilingual retrieval setting. Our best baseline achieves an average of 18.7 F1 points on \xorfull.
This result indicates that
\tydixor poses unique challenges  to tackle toward building a real-world open-retrieval QA system for diverse languages. 
We expect that our  dataset opens up  new challenges  to make progress in multilingual representation learning.

\section{The \tydixor~Dataset}\begin{figure*}[t!]
\vspace{-0.5em}
  \includegraphics[width=\linewidth]{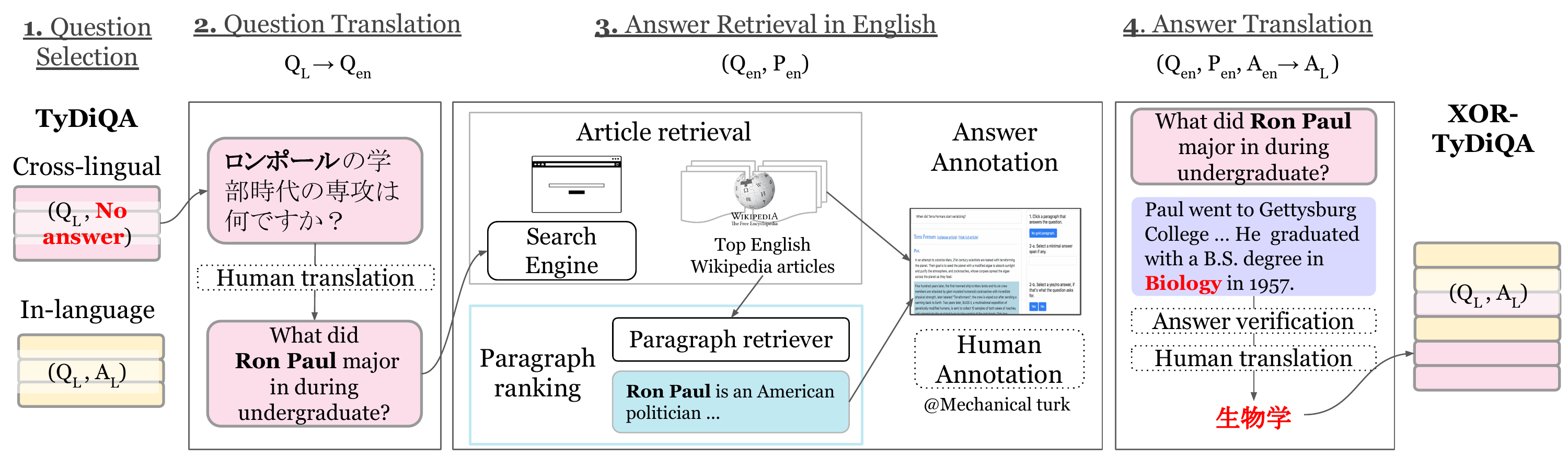}
\vspace{-0.5em}
  \caption{Overview of the annotation process for \tydixor. }
\vspace{-0.5em}
  \label{img:overview_annotation}
\end{figure*}
Our \tydixor dataset comprises questions inherited from \tydi~\cite{tydiqa} and answers augmented with our annotation process across 7 typologically diverse languages. We focus on cross-lingual retrieval from English Wikipedia because in our preliminary investigation we were able to find answers to a majority of the questions from resource-rich English Wikipedia, and native speakers with much annotation experience were readily available via crowdsourcing in English.
\label{sec:dataset}

\subsection{\tydixor Collection}
Our annotation pipeline proceeds with four steps: 1) collection of questions from \tydi without a same-language answer which require cross-lingual reference to answer (\S\ref{sec:question_collection}); 2) question translation from a target language to the pivot language of English where the missing information may exist (\S\ref{sec:question_trans_dataset}); 3) answer retrieval in the pivot language given a set of candidate documents (\S\ref{sec:qa_annot_dataset}); 4) answer verification and translation from the pivot language back to the original language (\S\ref{sec:answer_trans_dataset}).
Fig.~\ref{img:overview_annotation} shows an overview of the pipeline.

\subsubsection{Question Selection}
\label{sec:question_collection}
Our questions are collected from {\it unanswerable} questions in \tydi. 
A question is unanswerable in \tydi if an annotator cannot select a passage answer (a paragraph in the article that contains an answer).
We randomly sample 5,000 questions without any passage answer annotations (unanswerable questions) from the \tydi~training data, and split them into training (4,500) and development (500) sets. We use the development data from \tydi as our test data, since the \tydi's original test data is not publicly available.\footnote{Furthermore, despite the benefits of hidden test sets, the resource-intensive nature of open-retrieval QA is not suitable to code-submission leaderboards. This further precluded the use of the original \tydi~test sets.}
We choose 7 languages with varying amounts of Wikipedia data out of the 10 non-English languages based on the cost and availability of translators:\footnote{The cost of translations depends on the number of available translators, and the estimated translation cost for the other three non-English languages was considerably higher.} Arabic, Bengali, Finnish, Japanese, Korean, Russian and Telugu.

\subsubsection{Question Translation} 
\label{sec:question_trans_dataset}
We use a professional translation service, Gengo,\footnote{\url{https://gengo.com/}.} to translate all collected questions into English. Since named entities are crucial for QA, we instruct translators to carefully translate them by searching for common English translations from English Wikipedia or other external sources.
We perform manual quality assessment by native speakers on 50 translation samples, finding that more than 95\% are correct.
Note that while these translations are a part of the annotation procedure (due to the inherently cross-lingual nature of this task), they are \textit{not} provided to models during evaluation.

\subsubsection{Answer Retrieval in English}
\label{sec:qa_annot_dataset}
We use Amazon Mechanical Turk to retrieve answers to translated English questions given English Wikipedia articles. Annotators are instructed to select passage answers (gold paragraphs) and minimal answer spans as in \citet{tydiqa}. 

To annotate answers to information-seeking queries, previous work first identifies relevant Wikipedia articles using Google Search, and then annotators attempt to find answers there. 
\citet{asai2020unanswerable} show that in information-seeking QA datasets many questions were annotated as ``unanswerable'' due to two systematic errors: {\it retrieval error} where the search engine failed to retrieve a relevant article and {\it answer annotation error} where the annotator overlooks answer content.
Importantly, these two types of annotation errors present a tradeoff: if we retrieve many articles, retrieval errors will be reduced at the expense of answer annotation errors because annotators have to find answer context among many candidate articles.

\vspace{0.1cm}
\noindent {\bf Collaborative model-in-the-loop.} 
To find a middle ground in the tradeoff, we introduce a collaborative model-in-the-loop framework that uses Google Search and a state-of-the-art paragraph ranker.
We first run Google Search to retrieve as many as top 10 Wikipedia articles, resulting in 387 paragraphs per question on average. We score them with Path Retriever~\cite{Asai2020Learning} and present the five highest scoring paragraphs.
Annotators are asked to skim these five paragraphs first; if they cannot find any answer content, they are asked to read the rest of the paragraphs, where the Wikipedia section headings guide their reading.  
To incentivize workers to find answers beyond the pre-selected ones, we carefully communicate with workers and send additional rewards to annotators who actively read the rest of the paragraphs and find answers for questions that other annotators may overlook. We found about 70\% of the answers from the 5 paragraphs and 30\% from the rest of the paragraphs in the top 10 articles.
This means that while our paragraph ranking was effective, the annotators did not fully rely on it, thereby mitigating the influence of the passage ranking model on the dataset.
See Appendix \S\ref{ap_sec:annot_interface} for annotation interface details. 

\vspace{0.1cm}
\noindent {\bf Quality control for QA annotation.}
We first recruit MTurkers with a high approval rate ($\geq$ 96\%) located in English-speaking countries, and all workers first annotate the same qualification batch.
We assess the quality of those submissions and select high-quality annotators.
Consequently, 40 out of more than 200 workers were qualified and 24 workers annotated most of our data.
More details are in Appendix \ref{ap_sec:qual_qa}.

\subsubsection{Answer Verification and Translation}
\label{sec:answer_trans_dataset}
We verify the annotated answers and translate those answers back to the target languages (cross-lingual data). 
Finally, we mix the annotated cross-lingual data with the same-language data from \tydi to reflect the actual question distributions from native speakers (in-language data).

\noindent {\bf Answer verification.} We trained undergraduate students who are native English speakers to verify the annotated paragraphs and short answers. 
Only 8\% of the answers were marked as incorrect through the verification phase and were later corrected by our pool of high-quality crowdworkers who yielded less than 1\% annotation error. 

\vspace{.1cm}\noindent {\bf Answer translation.} 
We again use Gengo to translate answers from English back to the original languages. We give translators further instructions to normalize answers such that they are consistent with answers in \tydi. For example, some languages use their own unique set of numerals rather than Arabic numerals to represent numeric answers (e.g., Bengali numerals, Chinese numerals in Japanese text). 
The details of the answer translation process are described in Appendix~\S\ref{ap_sec:answer_trans_instructions}.
Note that because of the cost of answer translations, we conduct this answer translation process for evaluation sets only.

\subsection{The \tydixor Corpus}
\label{sec:corpus}
\begin{table}[t!]
\addtolength{\tabcolsep}{-2.4pt}
\small
    \centering
    \begin{tabular}{l|ccccccccc}
\toprule
\% & {\bf Ar} & {\bf Bn} & {\bf Fi} & {\bf Ja} & {\bf Ko} & {\bf Ru} & {\bf Te} & All\\ \midrule
\tydi & 82& 42 & 57 & 50& 29 & 69 & 28 & 50 \\
\tydixor & 92 & 82 &  83 & 77 & 68 & 83 & 44 & 72 \\\hdashline
Improvement & 10 & 40 &  26 & 27 & 39 & 14 & 16 & 22 \\
 \bottomrule
    \end{tabular}
    \caption{
    Percentage of the questions with short answers (answerable questions) in the original \tydi dataset (dev) ~and \tydixor. 
     The third row (Improvement) represents the percentage of the questions that become answerable by searching the English Wikipedia articles.
    }
    \label{tab:answer_recall_tydi_xor}
\end{table}
\begin{table}[t!]
\addtolength{\tabcolsep}{-0.6pt}
\small
\centering
\begin{tabular}{l |rrr |rrr }
\toprule
 &\multicolumn{3}{c}{Cross-lingual} & \multicolumn{3}{|c}{In-language}\\
 & {\bf Train } & {\bf Dev } & {\bf Test } & {\bf Train } & {\bf Dev } & {\bf Test} \\
\midrule
Ar & 2,574 &  350  & 137    & 15,828 & 358 &  1,132   \\
Bn  &2,582  & 312  & 128 &  2,428 & 115 & 139   \\
Fi &2,088 &  360 & 530 & 7,680 & 255 & 1,197 \\
Ja & 2,288  & 296  & 449 &5,527 & 137 & 867 \\
Ko &2,469 & 299 &  646 & 1,856 & 72 & 505 \\
Ru &1,941 & 255  &  235  & 7,349 & 313 & 1,125\\
Te & 1,308 & 238 & 374 & 5,451 & 113 & 712 \\
 \bottomrule
    \end{tabular}
    \caption{
    Dataset size of the \tydixor corpus (answered data). 
    \textbf{Cross-lingual} data comes from our reannotated questions that did not originally have same-language answers in \tydi. \textbf{In-language} data are taken directly from answerable questions in \tydi.
    }
    \label{tab:stat_annotation}
\end{table}

\noindent {\bf Dataset statistics.}\footnote{
After our initial release in November 2020, we modified the \tydixor~data, and released a new version as \tydixor (v1.1). All results are based on v1.1.
}
Table~\ref{tab:answer_recall_tydi_xor} shows the percentages of the questions annotated with short answers in the original \tydi and our \tydixor, and Table~\ref{tab:stat_annotation} shows statistics of \tydixor. 
As seen in Table~\ref{tab:answer_recall_tydi_xor}, cross-lingual retrieval significantly increases the answer coverage in all languages by up to 40\% (Bengali), and consequently we found answers for more than 50\% of the original information-seeking questions in 6 out of the 7 languages.\footnote{We found in the Telugu data, certain types of questions are very frequent (e.g., what is the pin code of {\it X} mandal?).
Those questions often ask some specific information of local administration districts, and are often unanswerable because (a) they are typically not described in English Wikipedia and (b) the overall coverage of Telugu Wikipedia is quite low. 
}
This result confirms the effectiveness of searching multilingual document collections to improve the answer coverage. 
Detailed statistics of the numbers of long answers, short answers, and unanswered questions are in Appendix~\S\ref{ap_sec:full_data}.
We also release the 30k manually translated questions for our training set, which could be used to train multilingual models or machine translation models.

\vspace{.1cm}
\noindent {\bf Qualitative examples.}
\begin{table*}
\begin{center}
{\small
\begin{tabular}{ L{0.4cm} | L{4.0cm} | L{6.5cm}  | C{1.5cm}  | C{1.5cm} }
\toprule
$L$ & Original Question: $Q_L$ ($Q_{en}$) & Passage Answer: $P_{en} $  or $P_{L}$ & Minimal Answer in English: $A_{en}$ & Final Answer: $A_L$ \\
\midrule
Ko & \begin{CJK}{UTF8}{mj}1993년 프랑스 총리는 누구인가요? \end{CJK}(Who was the French Prime Minister in 1993?) &  Mayor of Neuilly-sur-Seine from 1983 to 2002, he was Minister of the Budget under Prime Minister Édouard Balladur (1993–1995). & Édouard Balladur & \begin{CJK}{UTF8}{mj}에두아르 발라뒤르 \end{CJK}\\\hline
Ru &  \begin{otherlanguage*}{russian}Какая средняя зарплата в Краснодаре на сегодняшний день?\end{otherlanguage*} (What is the average wage in \textbf{Krasnodar}?) & Krasnodar has the lowest unemployment rate among the cities of the Southern Federal District at 0.3\% of the total working-age population. In addition, Krasnodar holds the first place in terms of highest average salary---21,742 rubles per capita.
& 21,742 rubles & 21,742 \begin{otherlanguage*}{russian}рубля
\end{otherlanguage*}
 \\\hline
Ja & \begin{CJK}{UTF8}{min}{\bf 速水堅曹}はどこで製糸技術を学んだ？\end{CJK} (Where did {\bf Kenso Hayami} learn the silk-reeling technique?)  & \begin{CJK}{UTF8}{min}藩営前橋製糸所を前橋に開設。カスパル・ミュラーから直接、器械製糸技術を学び (he founded Hanei Maebashi Silk Mill and learned instrumental silk reeling techniques directly from Caspal Müller)\end{CJK} & -- & \begin{CJK}{UTF8}{min} 藩営前橋製糸所\end{CJK} (Hanei Maebashi Silk Mill)\\
\bottomrule
\end{tabular}
}
\end{center}
\caption{\label{tab:odataset_example} Examples newly annotated for Korean (Ko) and Russian (Ru) questions. The bottom example is an answerable question from \tydi for which only Japanese Wikipedia includes the correct answer.
}
\end{table*}
 Table~\ref{tab:odataset_example} illustrates that finding relevant articles from multilingual document collections is important to answer questions asked by users with diverse linguistic and cultural backgrounds. 
The first question is unanswerable in Korean Wikipedia, but there is a clear description about who was the prime minister of France at the time in English Wikipedia. 
The second example shows English Wikipedia sometimes contains rich information about a target language-specific topic (e.g., economy in Krasnodar, a city in Russia).
Those examples demonstrate the effectiveness of searching for answers in another language with more abundant knowledge sources.
In the last question of Table~\ref{tab:odataset_example}, on the other hand, only the Wikipedia of the target language can provide the answer. \xor allows for both retrieval paths. 

\vspace{.1cm}
\noindent {\bf Comparison with other datasets.} 
\begin{table}[ht!]
\small
\centering
\addtolength{\tabcolsep}{-1.2pt}
\begin{tabular}{p{0.89in}|p{0.62in}p{0.52in}p{0.5in}}
\toprule
\textbf{Dataset} & \textbf{Asked by native speakers} & \textbf{Open-retrieval} & \textbf{Cross-lingual}  \\
\midrule
\tydi          & \cmark & \xmark & \xmark \\
MLQA             &\xmark &\xmark  &  \cmark \\
XQuAD            & \xmark & \xmark & \xmark \\
MKQA             & \xmark & WikiData &  \xmark  \\
MLQA-R & \xmark & 21k sents & \cmark  \\
XQuAD-R &\xmark  & 13k sents &  \cmark \\
\midrule
\tydixor &  \cmark & Wikipedia &\cmark \\ 
\bottomrule
\end{tabular}
\caption{Comparison with recent multilingual QA datasets. MKQA's answers are aligned to WikiData.
}
\label{tab:compare_datasets}
\end{table}
Table \ref{tab:compare_datasets} compares \tydixor and existing multilingual QA datasets.
\tydixor has three key properties that are distinct from these QA benchmarks. 
First, since all questions are inherited from \tydi, they are information-seeking questions written by native speakers, and better reflect native speakers' interests and their own linguistic phenomena. 
This distinguishes \tydixor from translation-based datasets such as MLQA~\cite{lewis2019mlqa} and MKQA~\cite{mkqa}.
Second, our dataset requires cross-lingual retrieval unlike other multilingual datasets such as \tydi or XQuAD \cite{Artetxe:etal:2019}, which focus on same-language QA. 
Lastly, questions in \tydixor require open retrieval from Wikipedia, whereas MLQA-R and XQuAD-R~\cite{roy2020lareqa} limit the search space to matching each question with the predetermined 21k/31k sentences.

\section{\xor~Tasks and Baselines}\label{sec:task}
\begin{figure*}[t!]
  \includegraphics[width=\linewidth]{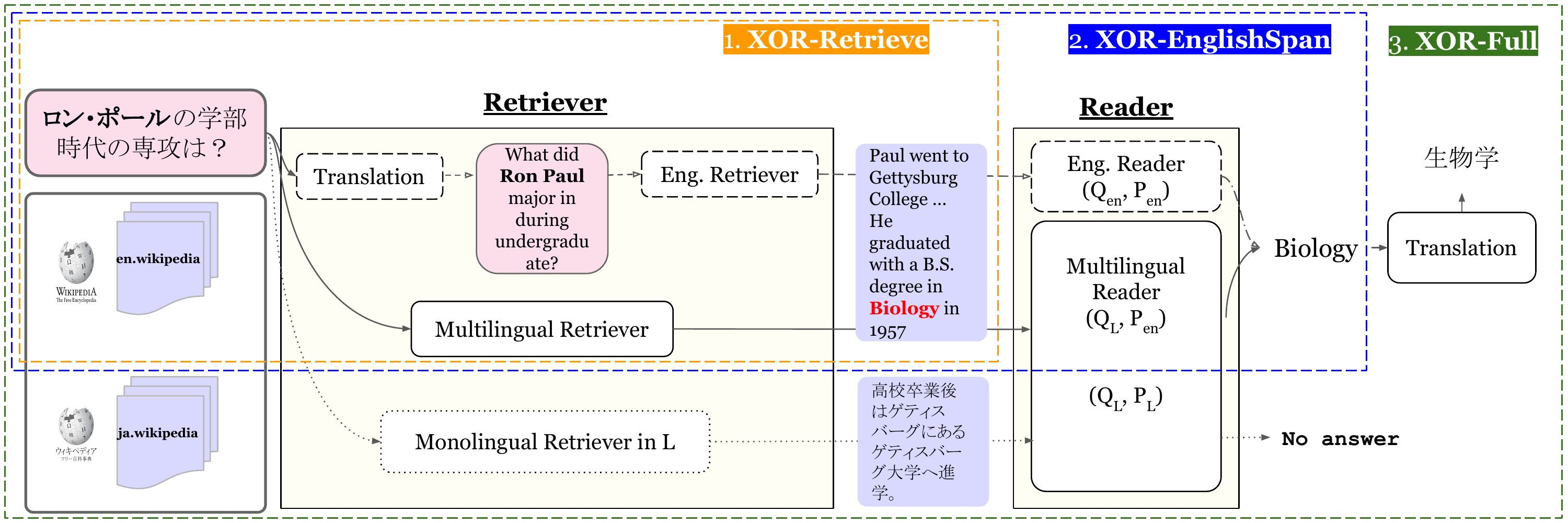}
  \caption{Overview of the tasks and baselines. Each dotted rectangle represents one of the three tasks and surrounds used pipeline modules.
  }
  \label{img:baseline_overview}
\end{figure*}
We introduce three new tasks (Fig.\ \ref{img:baseline_overview}): \textcolor{orange}{\xorretrieve}, \textcolor{blue}{\xorengspan}, and \textcolor{darkgreen}{\xorfull} with our newly collected \tydixor dataset and construct strong baselines for each task.
\xorfull defines our goal of building a multilingual open-retrieval QA system that uses both cross-lingual and in-language questions from \tydixor.
To diagnose where models fail and to allow researchers to use the data with less coding effort or computational resource, we also introduce the first two intermediate tasks that only use the cross-lingual data (Table \ref{tab:stat_annotation}).
We denote the target language by $L_i$. We also denote the English Wikipedia collection by $W_{eng}$ and the Wikipedia collection in each target language $L_i$ by $W_i$.
We experiment with baselines using black-box APIs as a reference, but we encourage the community to use white-box systems so that all experimental details can be understood. Nonetheless, we release the intermediate results from those external APIs to make our results reproducible. All of the white-box system results can be reproduced using our codebase. 

\subsection{\xorretrieve: Cross-lingual Paragraph Retrieval}
\label{sec:sub_task_retrieval}
\noindent {\bf Task.} Given a question in $L_i$ and English Wikipedia $W_{eng}$, the task is to retrieve English paragraphs for the question.
Finding evidence paragraphs from large-scale document collections like Wikipedia is a challenging task, especially when a query and documents are in different languages and systems cannot perform lexical matching.

\vspace{.1cm}
\noindent {\bf Evaluation.}
Different open-retrieval QA models use different units for retrieval.
To make fair comparisons across various models, we measure the recall by computing the fraction of the questions for which the minimal answer is contained in the top $n$ tokens selected. We evaluate with $n=2k, 5k$: R@2kt and R@5kt (kilo-tokens).

\vspace{.1cm} 
\noindent {\bf Translate baselines.} 
We first translate queries into English, and then paragraphs are retrieved in a monolingual way.
For query translation, we train transformer machine translation (MT) models on publicly available corpora for easy replication.
We also run Google's online machine translation service (GMT). This is not completely reproducible as these systems get constantly updated; nor do we know what model and training data they use.
We encourage the community to use open MT systems where system details are available.
For retrieval, we explore term-based retrieval (BM25, \citealt{10.1561/1500000019}), term-based retrieval followed by neural paragraph ranking (Path Retriever, \citealt{Asai2020Learning}), and end-to-end neural retrieval (DPR, \citealt{karpukhin2020dense}).

\vspace{.1cm}
\noindent {\bf Multilingual baselines.} Alternatively, we can directly apply a multilingual pretrained model to retrieve paragraphs.
We initialize and train a DPR encoder with multilingual BERT to enable multilingual document retrieval~\cite{devlin2018bert}.

\subsection{\xorengspan: L-to-English Open-Retrieval QA}
\label{sec:sub_task_definitions_1}
\noindent {\bf Task.}
Given a question in $L_i$ and English Wikipedia $W_{eng}$, a system retrieves paragraphs from $W_{eng}$ and extracts an answer.
This task is equivalent to existing open-retrieval QA tasks~\cite{chen2017reading}, except that the query is not in English.
This task involves challenging cross-lingual retrieval and question answering on the $L_i$ query and English evidence paragraphs.

\vspace{.1cm}
\noindent {\bf Evaluation.}
We use Exact Match (EM) and F1 over the annotated answer's token set following prior work~\cite{rajpurkar2016squad}.

\vspace{.1cm}
\noindent {\bf Baselines.}
Our pipeline uses a machine reading model to find a minimal span that answers the question given paragraphs selected from the previous \textsc{XOR-Retrieve} step. In particular, for the translate baselines, we use the same approach as state-of-the-art models \cite{Asai2020Learning,karpukhin2020dense} that jointly predicts a span and a relevance score of each paragraph to the question.
For the multilingual baseline where queries are \textit{not} automatically translated during evaluation, we build a reader model with multilingual BERT.

\vspace{-0.18cm}
\subsection{\xorfull: Round Trip}
\label{sec:main_task_definition}
\vspace{.1cm}
\noindent {\bf Task.}\hspace{1.5pt}
Given a question in target language $L_i$ and Wikipedia in both English and $L_i$ ($W_{eng}$ and $W_{i}$), a system is required to generate an answer in $L_i$.
In this task, a system does not know \textit{a priori} in which language we can find information that the user is seeking.
Note that the \xorfull~evaluation data includes both cross-lingual and in-language data, while \xorretrieve~and \xorengspan only use cross-lingual data during evaluation. 

\vspace{.1cm}
\noindent {\bf Evaluation.}\hspace{1.5pt}
Some answers in \xorfull are translated from English so the same spans may not exist in the target language's Wikipedia.
For this reason, we use token-level BLEU scores \cite{Papineni2001BleuAM} over a ground-truth token set in addition to F1 and EM. 
The same tokenizer is applied to ground-truth and predicted answers to compute token-level F1 and BLEU.\footnote{We use the Moses tokenizer \cite{koehn-etal-2007-moses} for all languages except we apply MeCab \cite{Kudo2005MeCabY} to Japanese.}

\vspace{.1cm}
\noindent {\bf Baselines.}\hspace{1pt}
Unlike the previous two tasks, evidence paragraphs can be found both in the target language and English, and a system has to output final answers based on the most plausible paragraphs. 
In this work, we introduce a simple multilingual baseline that first looks for answers in the target language and then English if no answers are found in the target language.
Specifically, we apply monolingual retrieval (i.e., BM25, Google Custom Search) for $W_i$ and a multilingual machine reading model based on XLM-RoBERTa~\cite{conneau-etal-2020-unsupervised} to find in-language answers in the target language (monolingual model; the bottom half of Fig.\ \ref{img:baseline_overview}).
If no answers are found by the monolingual model, we apply an \xorengspan baseline and translate English answers into the target language (the top half of Fig.\ \ref{img:baseline_overview}).

\section{Experiments and Analysis}

We present results from the baselines discussed above. We find that the three \xor tasks present challenges even for the strong models.
\subsection{Experimental Setup}
For training, we first finetune the retrieval and machine reading models with the Natural Questions data~\cite{kwiatkowski2019natural}  and then further finetune on our \tydixor~data.
For the BM25 retrieval baseline, we use ElasticSearch\footnote{\url{https://www.elastic.co/jp/}.} to store and search documents using BM25 similarities. 
For both Path Retriever and DPR, we run the official open-source code.
For our MT systems, we train base-sized (large for Russian) autoregressive transformers \cite{vaswani2017attention} on parallel corpora from OPUS~\cite{tiedemann-nygaard-2004-opus}, MultiUN \cite{ziemski-etal-2016-united}, or WMT19 \cite{barrault-etal-2019-findings}.
All data are encoded into subwords by BPE \cite{sennrich-etal-2016-neural} or SentencePiece \cite{kudo-richardson-2018-sentencepiece}.
We use the fairseq library \cite{ott-etal-2019-fairseq}.
Additional experimental details and full lists of hyperparameteres are available in Appendix \S\ref{ap_sec:training_details}.

We only evaluate questions having answers and do not give credit to predicting ``no answers'' as in prior open-retrieval work \cite{lee-chang-toutanova:2019:ACL2019}.
For \xorretrieve and \xorengspan,
we use cross-lingual data only and both cross-lingual and in-language data for \xorfull.
\subsection{\xorretrieve Experiments}Table~\ref{tab:main_results_retr} shows the R@5kt (as defined in \S\ref{sec:sub_task_retrieval}) for different retrieval and query translation systems.\footnote{We measured R@2kt as well (Table \ref{tab:main_results_retr_2k} in Appendix), but the relative pattern persisted across languages and methods.}
We also report the performance with the human English translations of the questions used during the dataset collection as an upper bound of translate baselines.  
The best R@5kt macro-averaged over the 7 languages comes from running DPR on human translations: 72.1. 
Machine translation systems achieve averages of 67.2 (GMT) and 50.0 (our MT) again with DPR.
The discrepancy between human and machine translation suggests that even state-of-the-art translation systems struggle to translate questions precisely enough to retrieve an evidence paragraph. 
Although the difference between GMT and our MT systems shows the effectiveness of industrial MT systems (large parallel data, model architecture, etc.),
there remains a substantial performance gap from human translation.
The translate baselines outperform the multilingual approach apart from Telugu, where our MT suffers from small parallel data (114k sentences), and as a result the multilingual approach performs better. 

BM25 substantially underperforms the other two models across the board.
DPR generally achieves similar performance, if not better, compared to Path Retriever despite the fact that Path Retriever was used in our annotation (\S \ref{sec:qa_annot_dataset}).
As we found that these patterns persisted in all the following experiments, we will only report results with DPR. 
\begin{table}[t!]
\small
\addtolength{\tabcolsep}{-3.0pt}

    \centering
    \begin{tabular}{l|ccc|cc|cc|c}
\toprule
 & \multicolumn{3}{c|}{Human Translation} &  \multicolumn{2}{c|}{GMT}  &  \multicolumn{2}{c|}{Our MT} & \textit{Multi.} \\
& DPR & \textsc{Path} & BM &  DPR  & \textsc{Path} & DPR  & \textsc{Path} & DPR \ \\ \midrule
Ar & 68.3 &  \textbf{70.0} &  41.6 & 67.5 &  63.3 & 52.5 & 51.6 & 50.4 \\
Bn & \textbf{85.6} & 82.0 & 57.0 & 83.2 & 78.9 & 63.2 & 64.8 & 57.7 \\
Fi & \textbf{73.1} & 70.2 & 43.7 & 68.1 & 64.1 & 65.9 & 59.5 & 58.9 \\
Ja & \textbf{68.9} & 63.0 & 38.8 & 60.1 & 52.3  & 52.1 & 41.7 & 37.3 \\
Ko & \textbf{70.9}& 63.6 & 43.8 & 66.3 &54.0 & 46.5 & 37.6 & 42.8  \\
Ru & \textbf{65.2}& 63.7 & 35.2 & 60.4 & 56.5 & 47.3 & 38.1 & 44.0 \\
Te & \textbf{72.2} & 64.1 & 44.6 & 65.0 & 62.5 & 22.7 & 18.1 & 44.9 \\
\midrule
Av.\ & \textbf{72.1} & 68.1 & 43.5 & 67.2 & 61.7 & 50.0 & 44.5 & 48.0 \\
 \bottomrule
    \end{tabular}
    \caption{R@5kt (\S\ref{sec:sub_task_retrieval}) on the test data in the \xorretrieve setting. \textsc{Path} and BM denote Path Retriever and BM25 respectively.
   {\it Multi.}~is a multilingual approach that bypasses the query translation step.
    }
    \label{tab:main_results_retr}
\end{table}

\vspace{-0.25cm}
\subsection{\xorengspan Experiments}\begin{table}[t!]
\small
\addtolength{\tabcolsep}{-1.5pt}

    \centering
    \begin{tabular}{l|cc|cc|cc|cc}
\toprule
& \multicolumn{2}{c|}{Human}  & \multicolumn{2}{c|}{GMT}  &  \multicolumn{2}{c|}{Our } & \multicolumn{2}{c}{\textit{Multi.}}\ \\
 & \multicolumn{2}{c|}{Translation } &  \multicolumn{2}{c|}{}  &  \multicolumn{2}{c|}{ MT} & \multicolumn{2}{c}{\textit{}}\ \\
& F1 & EM & F1 & EM  & F1 & EM & F1 & EM \\ \midrule
Ar & \textbf{43.2} & \textbf{32.8} & 39.5 & 28.5 & 28.0 & 23.4 & 17.9 & 11.7  \\
Bn &  \textbf{43.4} & \textbf{35.9} & 42.1 & 34.4 & 25.6 & 20.3 & 19.4 & 14.1 \\
Fi &  \textbf{34.8} & \textbf{26.0} & 28.2 & 21.3 & 29.3 & 22.1 & 24.5 & 18.3 \\
Ja & \textbf{29.9} & \textbf{22.3} & 23.5 & 17.4& 19.2 & 13.8  & 13.1 & 10.7 \\
Ko & \textbf{36.9} & \textbf{28.8} & 30.5 & 23.8 & 19.4& 14.2 & 14.3 & 9.9\\
Ru & \textbf{37.0} & \textbf{29.4} & 34.8 & 26.4 & 18.4 & 13.6 & 17.2 & 11.1 \\
Te & \textbf{42.4} & \textbf{35.0} & 31.6 & 25.1 & 3.8 & 2.7 & 14.4 & 10.2 \\
\midrule
Av.\ & \textbf{38.2} &\textbf{30.0} & 32.9 & 25.3 & 20.5 & 15.7 &17.2 & 12.3 \\
 \bottomrule
 \end{tabular}
    \caption{Performance on \xorengspan. The rightmost \textit{Multi.\ }section is a multilingual approach without query translation (\S\ref{sec:sub_task_retrieval}).
    }
    \label{tab:main_results_lte}
\end{table}
Table~\ref{tab:main_results_lte} shows the performance of the baseline models in \xorengspan.
The average macro F1 score with queries translated by human translators is 38.2, substantially higher than that of MT-based models: 32.9 and 20.5 F1 points for GMT and our MT respectively.
This suggests that errors in automatic query translation affect later layers in the pipeline.
The multilingual approach consistently underperforms translation-based methods, similarly to \xorretrieve.   
As in \xorretrieve, Telugu was an exception. The multilingual baseline significantly outperforms the translation-based approach with our MT system (14.4 vs.\ 3.6 F1 points). 
Query translation errors propagate to and directly impact downstream QA tasks in the languages with limited parallel data for MT training, and machine translation-based approaches may perform poorly. 
This encourages the research community to explore multilingual pretrained models to build a robust multilingual open-retrieval QA system for low-resource languages.

Similar to the original \tydi~dataset, the performance on \xorengspan varies across languages, which can be partially explained by the differing sets of questions \cite{tydiqa}.
The best baseline achieves 39.5 in Arabic compared to 23.5 F1 points in Japanese, which may come from differences in question difficulty as well as how the models are trained for each language. \label{sec:english_span}
\subsection{\xorfull Experiments}\begin{table*}[t!]
\addtolength{\tabcolsep}{-0.5pt}
\small
    \centering
    \begin{tabular}{ll|ll|ccccccc|ccc}
\toprule
\multicolumn{2}{c|}{Translation} &\multicolumn{2}{c|}{Retrieval} & \multicolumn{7}{c}{Target Language $L_i$ F1} & \multicolumn{3}{|c}{Macro Average}\\
Query & Answer & $L_i$ & Eng.\ & {\bf Ar} & {\bf Bn} & {\bf Fi} & {\bf Ja} & {\bf Ko} & {\bf Ru} & {\bf Te} & {\bf F1} & {\bf EM} & {\bf BLEU} \\ \midrule
 GMT & GMT & GS  & DPR &  \textbf{31.5} & 19.0 & \textbf{18.3} & \textbf{8.8} & \textbf{20.1} & \textbf{19.8} & \textbf{13.6} & \textbf{18.7} & \textbf{12.1} & \textbf{16.8} \\
Our MT & Our MT & GS & DPR & 29.6 & 6.6 & 15.5 & 7.6 & 16.4 & 18.7 & 1.7 & 13.7 & 8.7 & 12.0 \\
Our MT & Our MT &  BM25 & DPR & 12.1 & \textbf{22.0} & 9.3 & 5.4 & 9.7 & 7.4 & 0.8 & 9.5 & 6.0 & 8.9  \\
-- & GMT & GS & DPR & 30.5 & 10.6 & 16.9 & 8.2& 17.6 & \textbf{19.8} & 6.0  &15.7 & 10.0 & 13.9  \\
 \bottomrule
    \end{tabular}
    \caption{Performance on \xorfull (test data F1 scores). 
    ``GS'' denotes Google Search retrieval.
    }
    \label{tab:main_results_full}
\end{table*}
Table~\ref{tab:main_results_full} presents results on the \xorfull task. 
The first pipeline, which uses GMT, Google Search (GS), and DPR, yields the best average performance: 18.7 F1, 12.1 EM, and 16.8 BLEU points.
This indicates that systems like GMT and GS, which are typically trained on large data, are effective. Yet, we encourage the community to experiment on top of open systems such that all experimental details can be fully reported and understood. 
Replacing GMT with our MT (second row) results in a large performance drop in Bengali (6.6 vs.\ 19.0 F1 points) and Telugu (1.7 vs.\ 13.6). 
Further replacing GS with BM25 retrieval in the target languages (third row) causes a large performance drop in all languages (e.g., 9.7 vs.\ 16.4 in Korean). 
Consistent with the previous tasks, the multilingual approach shown in the forth row underperforms the translation-based counterpart (15.7 vs.\ 18.7 F1 points on average). 
Similar baselines perform considerably better in prior open-retrieval QA datasets, such as MKQA (30 EM points, \citealp{mkqa}) and NQ questions (40 F1, \citealp{karpukhin2020dense}). 
This gap illustrates the multidimensional challenge of \tydixor.

\subsection{Further Analysis}\begin{table*}[ht!]
\addtolength{\tabcolsep}{-0.45pt}
\small
    \centering
    \begin{tabular}{l|ccccccc| ccccccc}
\toprule
Query & \multicolumn{7}{c|}{MT BLEU} &  \multicolumn{7}{c}{\xorengspan F1} \\ 
Translator& Ar & Bn & Fi & Ja & Ko & Ru & Avg & Ar & Bn & Fi & Ja & Ko & Ru & Avg \\  \midrule
GMT &53.9 & 86.9 & 30.2 & 38.2 & 44.7 & 52.9& 51.8 &  35.4 & 42.1 & 31.8 & 27.2 & 32.5 & 34.7 & 34.0  \\
Our MT & 33.7 & 30.8 &  27.4 & 19.7 & 30.8 & 21.7 & 27.4 &  20.9 & 25.2 &  31.9 & 19.6  & 25.3  & 16.1 & 23.2  \\
Helsinki & 35.9 & 33.0  & 29.8 & 19.8  & 31.8 & 37.3 &  31.1 & 28.4& 21.3& 30.6& 19.0 &  25.3  & 29.6 & 25.7 \\
 \bottomrule
    \end{tabular}
    \caption{F1 scores on \xorengspan and the BLEU scores in query translation on the dev set. All configurations use DPR.
    Telugu is excluded since Helsinki does not support it as of October, 2020.
    }
    \label{tab:mt_performance}
\end{table*}

\vspace{.1cm}
\noindent {\bf Effects of translation performance on overall QA results.}
Table~\ref{tab:mt_performance} compares the query translation BLEU scores and the final QA F1 performance of the translation-based baseline with three different MT systems in \xorengspan: {\bf GMT}, {\bf Our MT}, and {\bf Helsinki} \cite{TiedemannThottingal:EAMT2020}.
GMT significantly outperforms the other two baselines, demonstrating that its training setup may yield large improvements in these languages; similarly, in cases where additional parallel training data is not available, multilingual models may remain strong modeling tools. 
On the other hand, it is noteworthy that high BLEU scores do not always lead to better QA performance.
In Bengali and Finnish, while Helsinki achieves a considerably better BLEU score than our MT (33.0 vs.\ 30.8 in Bengali and 29.8 vs.\ 27.4 in Finnish), our MT is 3.9 and 1.3 F1 points better in downstream \xorengspan, respectively.
See Appendix~\S\ref{ap_sec:trans_errors} for an example of translation errors resulting in QA errors.
Those results suggest that the BLEU score is not always indicative of the downstream performance and that evaluating MT performance in the context of \xor~would be important for improvements of multilingual QA systems. 

\vspace{.1cm}
\noindent {\bf Single language Wikipedia ablations in \xorfull.}
To assess our models' ability to benefit from multilingual collections, we try restricting the retrieval target to single language Wikipedia: English $W_{eng}$ only or target language $W_{i}$ only.
In $W_{eng}$ only, the best system, which applies GMT and DPR, underperforms the best pipeline that uses both $\mathbf{W}_{i, eng}$ in all languages except for Finnish and Japanese.
Similarly, the $W_{i}$ only setting generally underperforms the best $\mathbf{W}_{i, eng}$ pipeline.
These results illustrate the importance of searching multilingual collections.
See Table \ref{tab:main_results_full_f1_all} for the full results.
\section{Related Work}\label{sec:related}
\noindent {\bf Multilingual QA}
Much recent effort has been made to create non-English QA datasets to overcome the data scarcity in non-English languages. 
In addition to the datasets we already discussed in \S\ref{sec:corpus}, several other non-English reading comprehension datasets have been created~\cite{asai2018multilingual,lim2019korquad1,mozannar-etal-2019-neural,dhoffschmidt-etal-2020-fquad}.
\citet{liu-etal-2019-xqa} developed a template-based \textit{cloze} task, leading to different data distributions from realistic questions with a great degree of lexical overlap between questions and reference paragraphs \cite{lee-chang-toutanova:2019:ACL2019}.
More recently, \citet{hardalov-etal-2020-exams} introduced EXAMS, a multilingual multiple-choice reading comprehension dataset from school exams.

Our \tydixor is also closely related to QA@CLEF 2003-2008~\cite{magnini2003multiple,magnini2004overview,vallin2005overview,magnini2006overview,giampiccolo2007overview,forner2008overview}; both QA@CLEF and \tydixor attempt to develop and evaluate multilingual QA systems. Nevertheless,
there are three crucial differences.
First, our \tydixor has a large number of questions that are required for training current state-of-the-art QA models like DPR, while QA@CLEF only has 200 evaluation questions for each language without training data ~\cite{forner-etal-2010-evaluating}.
Secondly, the languages tested in QA@CLEF are all European languages, with the one exception of Indonesian; \tydixor includes typologically diverse languages.
Lastly, the task setup of QA@CLEF 2003-2008 is either monolingual---questions and documents are written in the same non-English language---or cross-lingual---the source and target languages are pre-specified~\cite{forner-etal-2010-evaluating}. In XOR QA, questions are asked in a target language but a system does not know in which language it can find an answer in a non-parallel Wikipedia collection. 
Those differences from QA@CLEF tasks better simulate real-world scenarios and introduce new challenges that have yet to be extensively studied. 

\vspace{.1cm}
\noindent {\bf Cross-lingual Information Retrieval}
Cross-lingual Information Retrieval (CLIR) is the task of retrieving relevant documents when the document collection is in a different language from the query language \cite{Hull1996QueryingAL}.
The retrieval component in \xor is closely related to CLIR, but differs in several critical ways.
First, since the end goal of \xor is QA, \xor queries always take question forms rather than search key words.
Further, while CLIR typically retrieves documents from a single (low-resource) language \cite{Zhang2019ImprovingLC}, \xor considers documents from both English and the query language.
In many applications, we do not know \textit{a priori} in which language we can find target information.
Lastly, our document collection is orders of magnitude bigger than typical CLIR benchmarks \cite{Sasaki2018CrossLingualLW, Zhang2019ImprovingLC}.

\section{Conclusion}We presented the task of \xor, in which a system retrieves and reads documents across languages to answer non-English information-seeking questions. 
We introduced a new large-scale \xor~dataset, \tydixor, with 40k newly annotated open-retrieval questions that cover seven typologically diverse languages. 
Our experiments showed that \tydixor is a challenging benchmark that can benefit from further effort in both QA and multilinguality  communities.

\newpage
\section*{Acknowledgments}
This research was supported by gifts from Google, the Allen Distinguished Investigator Award, the Sloan Fellowship, and the Nakajima Foundation Fellowship. 
We thank Sewon Min, Kristina Toutanova, David Wadden, the members of the UW NLP group, and the anonymous reviewers for their insightful feedback on this paper, Nancy Li, Xun Cao, Hitesh Boinpally, Samek Mulepati, Casey Zhao, Vitaly Nikolaev, Soumyadip Sengupta, Bindita Chaudhuri, and Aditya Kusupati for their help on our annotations and dataset proofing, and Nelson Liu and Pradeep Dasigi for their suggestions on the annotation interface and Amazon Mechanical Turk crowdsourcing.

\section*{Legal and Ethical Considerations}\paragraph{Were workers told what the dataset would be
used for and did they consent?}
Crowdworkers consented to have their responses used in this way through the Amazon Mechanical Turk Participation Agreement.
\paragraph{If it relates to people, could this dataset expose people to harm or legal action?}
Our dataset can include incorrect information to the extent that Wikipedia can have wrong information about people.
Nonetheless, we performed extensive quality control and answer verification to minimize the risk of harming people.
\paragraph{If it relates to people, does it unfairly advantage
or disadvantage a particular social group?}
One fundamental problem with the existing question answering benchmarks is that most of their questions are written by native English speakers and overly represent English-centric topics, such as American politics, sports, and culture. 
As such, models trained and developed on those datasets are likely to fail to serve people with diverse language and cultural backgrounds.
\tydixor remedies this long-standing problem by annotating questions from native speakers of diverse languages. Thus, we encourage researchers and developers to benchmark on \tydixor to mitigate the potential bias and unfairness of QA systems. We acknowledge, however, that this dataset still covers a very limited subset of languages in the world. We release a datasheet~\cite{Gebru2018DatasheetsFD} for our dataset to further document ethical implications.\footnote{\url{https://nlp.cs.washington.edu/xorqa/XORQA_site/xorqa_datasheet.pdf}.}

\newpage
\bibliography{anthology,custom}

\begin{thebibliography}{46}
\expandafter\ifx\csname natexlab\endcsname\relax\def\natexlab#1{#1}\fi

\bibitem[{Artetxe et~al.(2020)Artetxe, Ruder, and Yogatama}]{Artetxe:etal:2019}
Mikel Artetxe, Sebastian Ruder, and Dani Yogatama. 2020.
\newblock \href {https://arxiv.org/abs/1910.11856} {On the cross-lingual
  transferability of monolingual representations}.
\newblock In \emph{ACL}.

\bibitem[{Asai and Choi(2020)}]{asai2020unanswerable}
Akari Asai and Eunsol Choi. 2020.
\newblock \href {https://arxiv.org/abs/2010.11915} {Challenges in information
  seeking {QA}: Unanswerable questions and paragraph retrieval}.

\bibitem[{Asai et~al.(2018)Asai, Eriguchi, Hashimoto, and
  Tsuruoka}]{asai2018multilingual}
Akari Asai, Akiko Eriguchi, Kazuma Hashimoto, and Yoshimasa Tsuruoka. 2018.
\newblock \href {https://arxiv.org/abs/1809.03275} {Multilingual extractive
  reading comprehension by runtime machine translation}.

\bibitem[{Asai et~al.(2020)Asai, Hashimoto, Hajishirzi, Socher, and
  Xiong}]{Asai2020Learning}
Akari Asai, Kazuma Hashimoto, Hannaneh Hajishirzi, Richard Socher, and Caiming
  Xiong. 2020.
\newblock \href {https://arxiv.org/abs/1911.10470} {Learning to retrieve
  reasoning paths over {Wikipedia} graph for question answering}.
\newblock In \emph{ICLR}.

\bibitem[{Barrault et~al.(2019)Barrault, Bojar, Costa-juss{\`a}, Federmann,
  Fishel, Graham, Haddow, Huck, Koehn, Malmasi, Monz, M{\"u}ller, Pal, Post,
  and Zampieri}]{barrault-etal-2019-findings}
Lo{\"\i}c Barrault, Ond{\v{r}}ej Bojar, Marta~R. Costa-juss{\`a}, Christian
  Federmann, Mark Fishel, Yvette Graham, Barry Haddow, Matthias Huck, Philipp
  Koehn, Shervin Malmasi, Christof Monz, Mathias M{\"u}ller, Santanu Pal, Matt
  Post, and Marcos Zampieri. 2019.
\newblock \href {https://www.aclweb.org/anthology/W19-5301} {Findings of the
  2019 conference on machine translation ({WMT}19)}.
\newblock In \emph{WMT}.

\bibitem[{Callahan and Herring(2011)}]{callahan2011cultural}
Ewa~S Callahan and Susan~C Herring. 2011.
\newblock \href {https://onlinelibrary.wiley.com/doi/abs/10.1002/asi.21577}
  {Cultural bias in {Wikipedia} content on famous persons}.
\newblock \emph{JASIST}.

\bibitem[{Chen et~al.(2017)Chen, Fisch, Weston, and Bordes}]{chen2017reading}
Danqi Chen, Adam Fisch, Jason Weston, and Antoine Bordes. 2017.
\newblock \href {https://www.aclweb.org/anthology/P17-1171} {Reading
  {Wikipedia} to answer open-domain questions}.
\newblock In \emph{ACL}.

\bibitem[{Chen and Yih(2020)}]{chen-yih-2020-open}
Danqi Chen and Wen-tau Yih. 2020.
\newblock \href {https://doi.org/10.18653/v1/2020.acl-tutorials.8} {Open-domain
  question answering}.
\newblock In \emph{ACL: Tutorial Abstracts}.

\bibitem[{Clark et~al.(2020)Clark, Choi, Collins, Garrette, Kwiatkowski,
  Nikolaev, and Palomaki}]{tydiqa}
Jonathan~H. Clark, Eunsol Choi, Michael Collins, Dan Garrette, Tom Kwiatkowski,
  Vitaly Nikolaev, and Jennimaria Palomaki. 2020.
\newblock \href {https://arxiv.org/abs/2003.05002} {{TyDi QA}: A benchmark for
  information-seeking question answering in typologically diverse languages}.
\newblock \emph{TACL}.

\bibitem[{Conneau et~al.(2020)Conneau, Khandelwal, Goyal, Chaudhary, Wenzek,
  Guzm{\'a}n, Grave, Ott, Zettlemoyer, and
  Stoyanov}]{conneau-etal-2020-unsupervised}
Alexis Conneau, Kartikay Khandelwal, Naman Goyal, Vishrav Chaudhary, Guillaume
  Wenzek, Francisco Guzm{\'a}n, Edouard Grave, Myle Ott, Luke Zettlemoyer, and
  Veselin Stoyanov. 2020.
\newblock \href {https://www.aclweb.org/anthology/2020.acl-main.747}
  {Unsupervised cross-lingual representation learning at scale}.
\newblock In \emph{ACL}.

\bibitem[{Devlin et~al.(2019)Devlin, Chang, Lee, and
  Toutanova}]{devlin2018bert}
Jacob Devlin, Ming-Wei Chang, Kenton Lee, and Kristina Toutanova. 2019.
\newblock \href {https://arxiv.org/abs/1810.04805} {{BERT}: Pre-training of
  deep bidirectional transformers for language understanding}.
\newblock In \emph{NAACL}.

\bibitem[{d{'}Hoffschmidt et~al.(2020)d{'}Hoffschmidt, Belblidia, Heinrich,
  Brendl{\'e}, and Vidal}]{dhoffschmidt-etal-2020-fquad}
Martin d{'}Hoffschmidt, Wacim Belblidia, Quentin Heinrich, Tom Brendl{\'e}, and
  Maxime Vidal. 2020.
\newblock \href {https://www.aclweb.org/anthology/2020.findings-emnlp.107}
  {{FQ}u{AD}: {F}rench question answering dataset}.
\newblock In \emph{Findings of EMNLP}.

\bibitem[{Forner et~al.(2010)Forner, Giampiccolo, Magnini, Pe{\~n}as, Rodrigo,
  and Sutcliffe}]{forner-etal-2010-evaluating}
Pamela Forner, Danilo Giampiccolo, Bernardo Magnini, Anselmo Pe{\~n}as,
  {\'A}lvaro Rodrigo, and Richard Sutcliffe. 2010.
\newblock \href {https://www.aclweb.org/anthology/L10-1320/} {Evaluating
  multilingual question answering systems at {CLEF}}.
\newblock In \emph{LREC}.

\bibitem[{Forner et~al.(2008)Forner, Pe{\~n}as, Agirre, Alegria, For{\u{a}}scu,
  Moreau, Osenova, Prokopidis, Rocha, Sacaleanu et~al.}]{forner2008overview}
Pamela Forner, Anselmo Pe{\~n}as, Eneko Agirre, I{\~n}aki Alegria, Corina
  For{\u{a}}scu, Nicolas Moreau, Petya Osenova, Prokopis Prokopidis, Paulo
  Rocha, Bogdan Sacaleanu, et~al. 2008.
\newblock \href
  {https://link.springer.com/chapter/10.1007/978-3-642-04447-2_34} {Overview of
  the {CLEF} 2008 multilingual question answering track}.
\newblock In \emph{CLEF}.

\bibitem[{Gebru et~al.(2018)Gebru, Morgenstern, Vecchione, Vaughan, Wallach,
  Daum{\'e}, and Crawford}]{Gebru2018DatasheetsFD}
Timnit Gebru, J.~Morgenstern, Briana Vecchione, Jennifer~Wortman Vaughan,
  H.~Wallach, Hal Daum{\'e}, and K.~Crawford. 2018.
\newblock \href
  {https://www.fatml.org/media/documents/datasheets_for_datasets.pdf}
  {Datasheets for datasets}.
\newblock In \emph{FAT/ML}.

\bibitem[{Giampiccolo et~al.(2007)Giampiccolo, Forner, Herrera, Pe{\~n}as,
  Ayache, Forascu, Jijkoun, Osenova, Rocha, Sacaleanu
  et~al.}]{giampiccolo2007overview}
Danilo Giampiccolo, Pamela Forner, Jes{\'u}s Herrera, Anselmo Pe{\~n}as,
  Christelle Ayache, Corina Forascu, Valentin Jijkoun, Petya Osenova, Paulo
  Rocha, Bogdan Sacaleanu, et~al. 2007.
\newblock \href
  {https://link.springer.com/chapter/10.1007/978-3-540-85760-0_27} {Overview of
  the {CLEF} 2007 multilingual question answering track}.
\newblock In \emph{CLEF}.

\bibitem[{Hardalov et~al.(2020)Hardalov, Mihaylov, Zlatkova, Dinkov, Koychev,
  and Nakov}]{hardalov-etal-2020-exams}
Momchil Hardalov, Todor Mihaylov, Dimitrina Zlatkova, Yoan Dinkov, Ivan
  Koychev, and Preslav Nakov. 2020.
\newblock \href {https://www.aclweb.org/anthology/2020.emnlp-main.438}
  {{EXAMS}: A multi-subject high school examinations dataset for cross-lingual
  and multilingual question answering}.
\newblock In \emph{EMNLP}.

\bibitem[{Hull and Grefenstette(1996)}]{Hull1996QueryingAL}
David~A. Hull and Gregory Grefenstette. 1996.
\newblock \href
  {https://www.researchgate.net/publication/221299073_Querying_Across_Languages_A_Dictionary-Based_Approach_to_Multilingual_Information_Retrieval}
  {Querying across languages: a dictionary-based approach to multilingual
  information retrieval}.
\newblock In \emph{SIGIR}.

\bibitem[{Karpukhin et~al.(2020)Karpukhin, O{\u{g}}uz, Min, Wu, Edunov, Chen,
  and Yih}]{karpukhin2020dense}
Vladimir Karpukhin, Barlas O{\u{g}}uz, Sewon Min, Ledell Wu, Sergey Edunov,
  Danqi Chen, and Wen-tau Yih. 2020.
\newblock \href {https://arxiv.org/abs/2004.04906} {Dense passage retrieval for
  open-domain question answering}.
\newblock In \emph{EMNLP}.

\bibitem[{Koehn et~al.(2007)Koehn, Hoang, Birch, Callison-Burch, Federico,
  Bertoldi, Cowan, Shen, Moran, Zens, Dyer, Bojar, Constantin, and
  Herbst}]{koehn-etal-2007-moses}
Philipp Koehn, Hieu Hoang, Alexandra Birch, Chris Callison-Burch, Marcello
  Federico, Nicola Bertoldi, Brooke Cowan, Wade Shen, Christine Moran, Richard
  Zens, Chris Dyer, Ond{\v{r}}ej Bojar, Alexandra Constantin, and Evan Herbst.
  2007.
\newblock \href {https://www.aclweb.org/anthology/P07-2045} {{M}oses: Open
  source toolkit for statistical machine translation}.
\newblock In \emph{ACL System Demonstrations}.

\bibitem[{Kudo(2006)}]{Kudo2005MeCabY}
Taku Kudo. 2006.
\newblock \href {https://taku910.github.io/mecab/} {{MeCab}: Yet another
  part-of-speech and morphological analyzer}.

\bibitem[{Kudo and Richardson(2018)}]{kudo-richardson-2018-sentencepiece}
Taku Kudo and John Richardson. 2018.
\newblock \href {https://www.aclweb.org/anthology/D18-2012} {{S}entence{P}iece:
  A simple and language independent subword tokenizer and detokenizer for
  neural text processing}.
\newblock In \emph{EMNLP System Demonstrations}.

\bibitem[{Kwiatkowski et~al.(2019)Kwiatkowski, Palomaki, Redfield, Collins,
  Parikh, Alberti, Epstein, Polosukhin, Devlin, Lee, Toutanova, Jones, Kelcey,
  Chang, Dai, Uszkoreit, Le, and Petrov}]{kwiatkowski2019natural}
Tom Kwiatkowski, Jennimaria Palomaki, Olivia Redfield, Michael Collins, Ankur
  Parikh, Chris Alberti, Danielle Epstein, Illia Polosukhin, Jacob Devlin,
  Kenton Lee, Kristina Toutanova, Llion Jones, Matthew Kelcey, Ming-Wei Chang,
  Andrew~M. Dai, Jakob Uszkoreit, Quoc Le, and Slav Petrov. 2019.
\newblock \href {https://www.aclweb.org/anthology/Q19-1026} {{Natural
  Questions}: A benchmark for question answering research}.
\newblock \emph{TACL}.

\bibitem[{Lee et~al.(2019)Lee, Chang, and
  Toutanova}]{lee-chang-toutanova:2019:ACL2019}
Kenton Lee, Ming-Wei Chang, and Kristina Toutanova. 2019.
\newblock \href {https://www.aclweb.org/anthology/P19-1612} {Latent retrieval
  for weakly supervised open domain question answering}.
\newblock In \emph{ACL}.

\bibitem[{Lewis et~al.(2020)Lewis, O\u{g}uz, Rinott, Riedel, and
  Schwenk}]{lewis2019mlqa}
Patrick Lewis, Barlas O\u{g}uz, Ruty Rinott, Sebastian Riedel, and Holger
  Schwenk. 2020.
\newblock \href {https://arxiv.org/abs/1910.07475} {{MLQA}: Evaluating
  cross-lingual extractive question answering}.
\newblock In \emph{ACL}.

\bibitem[{Lim et~al.(2019)Lim, Kim, and Lee}]{lim2019korquad1}
Seungyoung Lim, Myungji Kim, and Jooyoul Lee. 2019.
\newblock \href {https://arxiv.org/abs/1909.07005} {{KorQuaAD}1.0: Korean {QA}
  dataset for machine reading comprehension}.

\bibitem[{Liu et~al.(2019)Liu, Lin, Liu, and Sun}]{liu-etal-2019-xqa}
Jiahua Liu, Yankai Lin, Zhiyuan Liu, and Maosong Sun. 2019.
\newblock \href {https://www.aclweb.org/anthology/P19-1227} {{XQA}: A
  cross-lingual open-domain question answering dataset}.
\newblock In \emph{ACL}.

\bibitem[{Longpre et~al.(2020)Longpre, Lu, and Daiber}]{mkqa}
Shayne Longpre, Yi~Lu, and Joachim Daiber. 2020.
\newblock \href {https://arxiv.org/abs/2007.15207} {{MKQA}: A linguistically
  diverse benchmark for multilingual open domain question answering}.

\bibitem[{Magnini et~al.(2006)Magnini, Giampiccolo, Forner, Ayache, Jijkoun,
  Osenova, Penas, Rocha, Sacaleanu, and Sutcliffe}]{magnini2006overview}
Bernardo Magnini, Danilo Giampiccolo, Pamela Forner, Christelle Ayache,
  Valentin Jijkoun, Petya Osenova, Anselmo Penas, Paulo Rocha, Bogdan
  Sacaleanu, and Richard Sutcliffe. 2006.
\newblock \href
  {https://link.springer.com/chapter/10.1007/978-3-540-74999-8_31} {Overview of
  the {CLEF} 2006 multilingual question answering track}.
\newblock In \emph{CLEF}.

\bibitem[{Magnini et~al.(2003)Magnini, Romagnoli, Vallin, Herrera, Penas,
  Peinado, Verdejo, and de~Rijke}]{magnini2003multiple}
Bernardo Magnini, Simone Romagnoli, Alessandro Vallin, Jes{\'u}s Herrera,
  Anselmo Penas, V{\'\i}ctor Peinado, Felisa Verdejo, and Maarten de~Rijke.
  2003.
\newblock \href
  {https://link.springer.com/chapter/10.1007/978-3-540-30222-3_46} {The
  multiple language question answering track at {CLEF} 2003}.
\newblock In \emph{CLEF}.

\bibitem[{Magnini et~al.(2004)Magnini, Vallin, Ayache, Erbach, Pe{\~n}as,
  De~Rijke, Rocha, Simov, and Sutcliffe}]{magnini2004overview}
Bernardo Magnini, Alessandro Vallin, Christelle Ayache, Gregor Erbach, Anselmo
  Pe{\~n}as, Maarten De~Rijke, Paulo Rocha, Kiril Simov, and Richard Sutcliffe.
  2004.
\newblock \href {https://link.springer.com/chapter/10.1007/11519645_38}
  {Overview of the {CLEF} 2004 multilingual question answering track}.
\newblock In \emph{CLEF}.

\bibitem[{{Miniwatts Marketing Group}(2011)}]{internetstat}
{Miniwatts Marketing Group}. 2011.
\newblock \href {https://www.internetworldstats.com/stats7.htm} {Internet world
  stats: Usage and population statistics}.

\bibitem[{Mozannar et~al.(2019)Mozannar, Maamary, El~Hajal, and
  Hajj}]{mozannar-etal-2019-neural}
Hussein Mozannar, Elie Maamary, Karl El~Hajal, and Hazem Hajj. 2019.
\newblock \href {https://www.aclweb.org/anthology/W19-4612} {Neural {A}rabic
  question answering}.
\newblock In \emph{WANLP}.

\bibitem[{Ott et~al.(2019)Ott, Edunov, Baevski, Fan, Gross, Ng, Grangier, and
  Auli}]{ott-etal-2019-fairseq}
Myle Ott, Sergey Edunov, Alexei Baevski, Angela Fan, Sam Gross, Nathan Ng,
  David Grangier, and Michael Auli. 2019.
\newblock \href {https://arxiv.org/abs/1904.01038} {fairseq: A fast, extensible
  toolkit for sequence modeling}.
\newblock In \emph{NAACL System Demonstrations}.

\bibitem[{Papineni et~al.(2002)Papineni, Roukos, Ward, and
  Zhu}]{Papineni2001BleuAM}
Kishore Papineni, Salim Roukos, Todd Ward, and Wei-Jing Zhu. 2002.
\newblock \href {https://www.aclweb.org/anthology/P02-1040.pdf} {{BLEU}: a
  method for automatic evaluation of machine translation}.
\newblock In \emph{ACL}.

\bibitem[{Rajpurkar et~al.(2016)Rajpurkar, Zhang, Lopyrev, and
  Liang}]{rajpurkar2016squad}
Pranav Rajpurkar, Jian Zhang, Konstantin Lopyrev, and Percy Liang. 2016.
\newblock \href {https://arxiv.org/abs/1606.05250} {{SQ}u{AD}: 100,000+
  questions for machine comprehension of text}.
\newblock In \emph{EMNLP}.

\bibitem[{Robertson and Zaragoza(2009)}]{10.1561/1500000019}
Stephen Robertson and Hugo Zaragoza. 2009.
\newblock \href {https://doi.org/10.1561/1500000019} {The probabilistic
  relevance framework: {BM25} and beyond}.
\newblock \emph{Foundations and Trends in Information Retrieval}.

\bibitem[{Roy et~al.(2020)Roy, Constant, Al-Rfou, Barua, Phillips, and
  Yang}]{roy2020lareqa}
Uma Roy, Noah Constant, Rami Al-Rfou, Aditya Barua, Aaron Phillips, and Yinfei
  Yang. 2020.
\newblock \href {https://arxiv.org/abs/2004.05484} {{LAReQA}: Language-agnostic
  answer retrieval from a multilingual pool}.
\newblock In \emph{EMNLP}.

\bibitem[{Sasaki et~al.(2018)Sasaki, Sun, Schamoni, Duh, and
  Inui}]{Sasaki2018CrossLingualLW}
Shota Sasaki, Shuo Sun, Shigehiko Schamoni, Kevin Duh, and Kentaro Inui. 2018.
\newblock \href {https://www.aclweb.org/anthology/N18-2073/} {Cross-lingual
  learning-to-rank with shared representations}.
\newblock In \emph{NAACL}.

\bibitem[{Sennrich et~al.(2016)Sennrich, Haddow, and
  Birch}]{sennrich-etal-2016-neural}
Rico Sennrich, Barry Haddow, and Alexandra Birch. 2016.
\newblock \href {https://www.aclweb.org/anthology/P16-1162} {Neural machine
  translation of rare words with subword units}.
\newblock In \emph{ACL}.

\bibitem[{Tiedemann and Nygaard(2004)}]{tiedemann-nygaard-2004-opus}
J{\"o}rg Tiedemann and Lars Nygaard. 2004.
\newblock \href {http://www.lrec-conf.org/proceedings/lrec2004/pdf/320.pdf}
  {The {OPUS} corpus - parallel and free}.
\newblock In \emph{LREC}.

\bibitem[{Tiedemann and Thottingal(2020)}]{TiedemannThottingal:EAMT2020}
J{\"o}rg Tiedemann and Santhosh Thottingal. 2020.
\newblock \href {https://www.aclweb.org/anthology/2020.eamt-1.61} {{OPUS-MT}
  — {B}uilding open translation services for the {W}orld}.
\newblock In \emph{EAMT}.

\bibitem[{Vallin et~al.(2005)Vallin, Magnini, Giampiccolo, Aunimo, Ayache,
  Osenova, Pe{\~n}as, De~Rijke, Sacaleanu, Santos et~al.}]{vallin2005overview}
Alessandro Vallin, Bernardo Magnini, Danilo Giampiccolo, Lili Aunimo,
  Christelle Ayache, Petya Osenova, Anselmo Pe{\~n}as, Maarten De~Rijke, Bogdan
  Sacaleanu, Diana Santos, et~al. 2005.
\newblock \href {https://link.springer.com/chapter/10.1007/11878773_36}
  {Overview of the {CLEF} 2005 multilingual question answering track}.
\newblock In \emph{CLEF}.

\bibitem[{Vaswani et~al.(2017)Vaswani, Shazeer, Parmar, Uszkoreit, Jones,
  Gomez, Kaiser, and Polosukhin}]{vaswani2017attention}
Ashish Vaswani, Noam Shazeer, Niki Parmar, Jakob Uszkoreit, Llion Jones,
  Aidan~N Gomez, {\L}ukasz Kaiser, and Illia Polosukhin. 2017.
\newblock \href {https://arxiv.org/abs/1706.03762} {Attention is all you need}.
\newblock In \emph{NeurIPS}.

\bibitem[{Zhang et~al.(2019)Zhang, Westerfield, Shim, Bingham, Fabbri, Verma,
  Hu, and Radev}]{Zhang2019ImprovingLC}
Rui Zhang, Caitlin Westerfield, Sungrok Shim, Garrett Bingham, Alexander~R.
  Fabbri, Neha Verma, William~T Hu, and Dragomir~R. Radev. 2019.
\newblock \href {https://arxiv.org/abs/1906.03492} {Improving low-resource
  cross-lingual document retrieval by reranking with deep bilingual
  representations}.
\newblock In \emph{ACL}.

\bibitem[{Ziemski et~al.(2016)Ziemski, Junczys-Dowmunt, and
  Pouliquen}]{ziemski-etal-2016-united}
Micha{\l} Ziemski, Marcin Junczys-Dowmunt, and Bruno Pouliquen. 2016.
\newblock \href {https://www.aclweb.org/anthology/L16-1561} {The {United
  Nations} parallel corpus v1.0}.
\newblock In \emph{LREC}.

\end{thebibliography}
\bibliographystyle{acl_natbib}

\newpage
\appendix
\section*{Appendix}
\section{Spirit behind Annotation Interface Design}
\paragraph{Open-retrieval annotation desiderata.} 
Open-retrieval QA annotation comes with unique challenges. 
In article-oriented QA such as SQuAD \cite{rajpurkar2016squad}, all labels are with regard to a single document and a single human can indeed read the whole document.
In open-retrieval QA, answers can be retrieved from millions of documents. Because exhaustively reading so much content is impossible for humans, the notion of ``human performance'' must be reconsidered in this context. This is why we only evaluate questions having answers in the open-retrieval setting and discard those where no answer was found---it is difficult to prove an answer does not exist in the millions of documents.

\paragraph{Limits of traditional annotation.}
In addition to fundamental problems of information scarcity and asymmetry in multilingual QA, questions can be labeled as unanswerable simply because of annotation errors.
Annotation procedures for information-seeking QA data usually have each annotator read a single Wikipedia article retrieved by a search engine and label a correct answer span or label the question as not answered by the article \cite{kwiatkowski2019natural, tydiqa}.
In this procedure, the answer coverage is underestimated when the search engine fails to retrieve relevant articles (\textit{retrieval errors}) or the annotator overlooks answer content from the selected articles (\textit{answer annotation errors}, \citealp{asai2020unanswerable}).
Importantly, these two types of annotation errors present a tradeoff: if we retrieve many articles, retrieval errors will be reduced at the expense of answer annotation errors because annotators have to find answer context among many candidate articles. An annotation procedure that misses too many answers will lead to an artificially small dataset.

\section{Additional Details of Dataset Creation}
\subsection{Annotation Interface}
\label{ap_sec:annot_interface}
\begin{figure}[ht!]
  \includegraphics[width=0.9\linewidth]{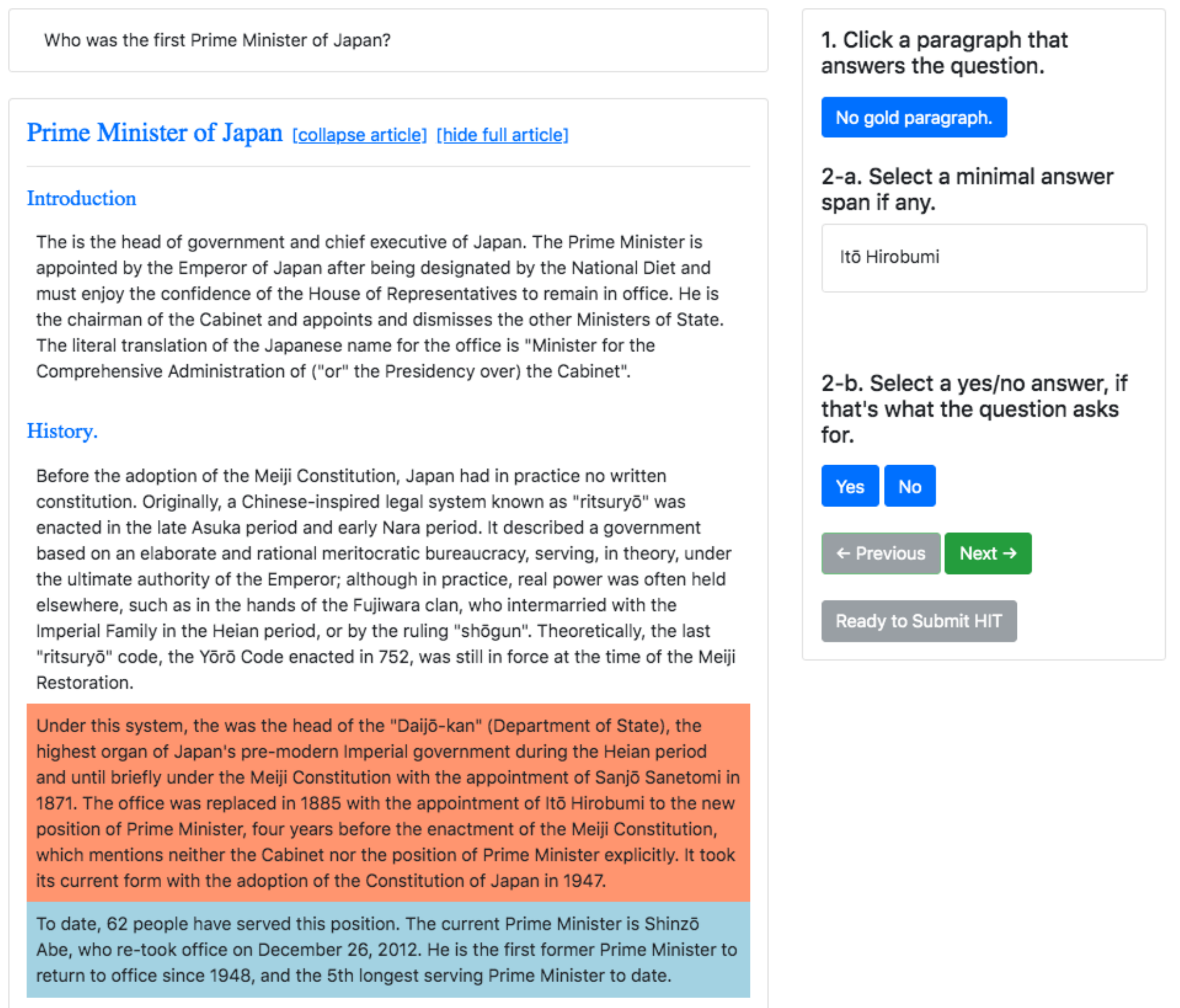}
  \caption{Annotation interface (expanded). The blue highlighted paragraphs are ranked high by the BERT paragraph ranker, and the orange highlighted paragraph is the one clicked by the annotator. }
  \label{img:annotation_ui_opened}
\end{figure}

\begin{figure}[ht!]
  \includegraphics[width=0.9\linewidth]{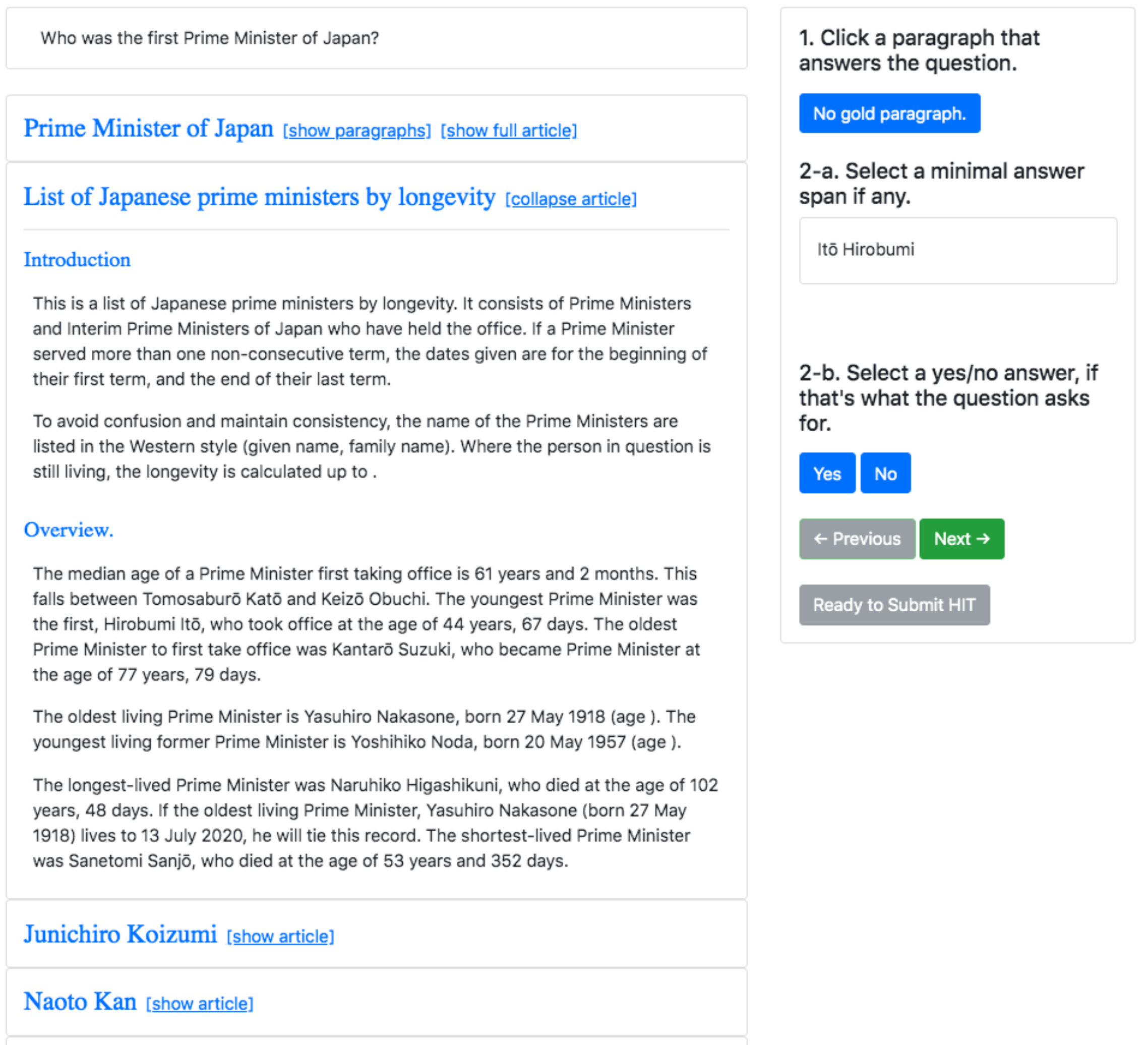}
  \caption{Annotation interface (collapsed). Annotators can choose to read full articles or collapse articles.}
  \label{img:annotation_ui_collapsed}
\end{figure}

In this section, we describe the details of the annotation interface we used for answer annotation in English (\S\ref{sec:qa_annot_dataset}). 
The annotation interface can be 
seen in Figs.\ \ref{img:annotation_ui_opened} and \ref{img:annotation_ui_collapsed}.
To maximize the answer coverage for open-retrieval questions, we first rank paragraphs from top articles retrieved by Google Search. 
During this paragraph ranking process, we only consider top 5 paragraphs and exclude the articles ranked from top 6 to 10. Increasing the number of the initial articles introduces more noise and confuses our paragraph ranking model, while human annotators sometimes found that those low-ranked articles relevant and retrieved answers from them as discussed in \S\ref{sec:qa_annot_dataset}. 
In the annotation interface, we first present those top 5 paragraphs first (the ones highlighted in light blue in Fig.~\ref{img:annotation_ui_opened}). 
When annotators do not find answers in the pre-selected top 5 paragraphs, they will explore more paragraphs and articles by expanding originally collapsed articles as in Fig.~\ref{img:annotation_ui_collapsed}. 


\subsection{Quality Control for Question Translation}
We first ask Gengo translators to translate 20 sample questions following our detailed instruction before starting the task, and ask native speakers to assess the quality of translations. 
We filter out translators who do not provide translation results that meet our standard (e.g., wrong translations of entities, heavy reliance on public machine translation systems).
We have found that some of the translators almost copy and paste outputs of existing APIs without fixing errors even when there are crucial errors. 
After this initial qualification process, we observe that the translation quality is sufficiently high.  

\subsection{Quality Control for QA annotation}
\label{ap_sec:qual_qa}
To control the QA annotation quality, we recruit workers with a high approval rate ($\geq 96\%$) located in English-speaking countries and conducted a rigorous qualification procedure. 
In our qualification stage, we post small calibration batches and evaluate the workers' performance by expert judgements from authors and agreement with other annotators. 
To keep the high quality of annotations, we randomly sample qualified workers weekly and manually monitor their annotations by comparing them with gold annotations by authors. 
We remove qualifications when we detect too many incorrect annotations (e.g., label a paragraph about a different person as a gold paragraph) and remove the annotations done by those disqualified annotators, which are later reannotated by a qualified worker. 
Over 200 annotators participated in our calibration tasks. 
About 40 workers are qualified with 24 actively working on the final dataset.
Each HIT contains 5 questions with a reward ranging  from 1.5 to 2.5 USD.
Qualified annotators generally spend 1-2 minutes to answer each question. 
We give special rewards to annotators who actively search additional paragraphs or articles; the amounts of the rewards are calculated based on the numbers of the HITs they have submitted, resulting in 5-10 USD for each payment. 

\subsection{Answer Translation Instructions}
\label{ap_sec:answer_trans_instructions}
During answer translation, we asked annotators to follow the instructions listed below: 
\vspace{-0.2cm}
\begin{itemize}
\itemsep-0.4em
    \item Translators need to use metric units by default, instead of imperial units. 
    \item If the original answers are expressed in an imperial unit, translators are encouraged to convert them into a metric unit (e.g., Height 5'3" --> \begin{CJK}{UTF8}{min}身長\end{CJK} 160 cm). 
    \item When translating proper nouns, translators are asked to use an official translation if it is available in Wikipedia; otherwise they are encouraged to transliterate them.
\end{itemize}
We also specify some language-specific instructions to make the translated answers consistent with the ones in the original \tydi dataset.
\vspace{-0.2cm}
\begin{itemize}
\itemsep-0.4em
    \item For Japanese and Korean, translators do not need to spell out the numbers (e.g., 1954 --> \begin{CJK}{UTF8}{min}千九百五十四\end{CJK}) as people usually use Arabic numerals. 
    \item For Bengali, we expect the numbers will be spelled out in Bengali numerals as Bengali speakers rarely use Arabic numerals.  
    \item For Japanese and Korean, translators use appropriate measure words (e.g., 1867\begin{CJK}{UTF8}{mj}년\end{CJK}, 57\begin{CJK}{UTF8}{min}歳\end{CJK}) if those measure words are commonly added in those languages.
    \item For the languages where the date needs to be expressed in some rigid format, translators need to follow the format.
\end{itemize}

\subsection{Full Data Statistics of Cross-lingual data}
\label{ap_sec:full_data}
Seen in Table \ref{tab:stat_annotation_incl_long} are full data statistics of cross-lingual data of \tydixor.
Among the questions with ``Long'' answer annotations are some questions without any short answers as in Natural Questions or \tydi. 
We do not include those ``Long answer only'' examples in our \tydixor~evaluations.
 \begin{table*}
\small
    \centering
    \begin{tabular}{l|ccc | ccc|ccc}
\toprule
{\bf $L_i$} & \multicolumn{3}{c}{{\bf Train (1 way)}} & \multicolumn{3}{c}{{\bf Dev (2 way)}} & \multicolumn{3}{c}{{\bf Test (2 way)}}  \\
 & Total & Long (\%) & Short (\%) &Total &Long (\%) & Short (\%) & total & Long (\%) & Short (\%) \\\midrule
Arabic & 4,500 & 2,862 (63) & 2,574 (57) & 500& 357 (71) & 350 (70) & 235 & 144 (61)& 137 (58) \\
Bengali & 4,500 & 2,822 (63) & 2,582 (57) & 500 & 330 (66) & 312 (62) & 185 & 131  (70) & 128 (69)  \\
Finnish & 4,500 & 2,454 (55) & 2,088 (46)& 500& 372 (74) & 360 (72)& 800& 556 (69) & 530 (66)  \\
Japanese & 4,500 & 2,557 (57) & 2,288 (51) & 500 & 320 (64) &  296 (60) & 779 & 477 (61)& 449 (58) \\
Korean & 4,500 & 2,674 (59) & 2,469 (55) & 500  & 314 (63) & 299 (60) & 1,177 & 684 (58) &  646 (55) \\
Russian & 4,500 &2,178 (48) &1,941  (43)  & 500 & 270 (54) & 255 (51) & 470&  252 (53)&  235 (50)  \\
Telugu & 4,500 &1,515 (33) & 1,308 (29)  & 500 & 258 (52) & 238 (47) & 1,752 & 394 (22) &  374 (21)  \\
 \bottomrule
    \end{tabular}
    \caption{Dataset statistics of the resulting \xor~corpus (cross-lingual data only). ``Long'' denotes the questions with paragraph answer annotations, and ``Short'' denotes the questions with short answer annotations. 
    During evaluation, we disregard the questions without short answer annotations.
    }
    \label{tab:stat_annotation_incl_long}
\end{table*}

\section{Training details}
\label{ap_sec:training_details}
We describe the details in training our baselines to facilitate easy replication of our results.
\subsection{Machine Translation Models}
Table \ref{mt-hyp} lists hyperpameters for training our transformer machine translation models. We generally follow the hyperprameters for the base-sized transformer \cite{vaswani2017attention}.
The one exception is English$\leftrightarrow$Russian where we used pretrained transformer large models.\footnote{\url{https://github.com/pytorch/fairseq/blob/master/examples/translation/README.md}.}
For each language direction, all data are encoded into subwords by Moses tokenization (\citealp{koehn-etal-2007-moses}, for Arabic, Finnish, and Russian) and BPE \cite{sennrich-etal-2016-neural} or SentencePiece (\citealp{kudo-richardson-2018-sentencepiece}, for Bengali, Japanese, Korean, and Telugu).
We train an autoregressive transformer \cite{vaswani2017attention} with the fairseq library \cite{ott-etal-2019-fairseq}.

\begin{table}[h]
\centering
\small
\begin{tabular}{@{} l@{\hspace{-0.2cm}} r @{}}
\toprule[.1em]
\textbf{Hyperparameter} & \textbf{Value}\\
\midrule[.1em]
label smoothing & 0.1\\
\# max tokens & 4096 \\
dropout rate & 0.3\\
encoder embedding dim  & 512\\
encoder ffn dim  & 2048\\
\# encoder attn heads & 8\\
decoder embedding dim  & 512\\
decoder ffn dim  & 2048\\
\# decoder attn heads & 8\\
max source positions & 10000 \\
max target positions & 10000 \\
Adam lrate& $5\times 10^{-4}$ \\
Adam $\beta_1$& 0.9\\
Adam $\beta_2$& 0.98\\
lr-scheduler &  inverse square \\
warm-up lr & $1\times 10^{-7}$ \\
\# warmup updates & 4000 \\
\# max updates &  300K \\
length penalty & 1.0\\
\bottomrule[.1em]
\end{tabular}
 \caption{Hyperparameters for our transformer machine translation models.}
 \label{mt-hyp}
\end{table}

\subsection{Retrieval Models}
\paragraph{Training DPR and Path Retriever.}
To train an English DPR and Path Retriever, we first initialize the parameters of the models with the ones trained on Natural Questions Open data, which is available on their repository. 
During finetuning on \tydixor, we use the human translated questions with the annotated gold paragraph data.

\paragraph{Choice of negative and positive context.}
Selection of positive and negative examples is crucial to train competitive neural retriever models~\cite{karpukhin2020dense}. 
We follow the hyperparameters used in the original papers \cite{karpukhin2020dense, Asai2020Learning}. 
To construct effective negative and positive context, we follow the approaches introduced by the authors of those works.  

To train DPR, we use the original gold paragraphs (long answers) annotated by MTurkers as positive passages.
Following the experimental settings of DPR on Natural Questions, we first split gold paragraphs into 100-token units, and consider the units with the original short answer annotations as positive context.
For negative context, we first randomly sample one negative paragraph per question from the top 5 paragraphs pre-selected by our paragraph reranking model in \S\ref{sec:qa_annot_dataset}, split the negative paragraph into 100-token units, and then randomly pick one to use it as a negative context.  
We also reuse the in-batch negative paragraphs as discussed in \citet{karpukhin2020dense}.

Regarding the training of Path Retriever, we randomly sample top 50 paragraphs from the top 10 articles retrieved for annotations and use them as negative paragraphs. 
We also use the annotated long answers as positive paragraphs.

\paragraph{Implementation details of BM25 Retrievers.}
To implement BM25-based retrievers for the 7 languages, we use ElasticSearch's Python client (Python Elasticsearch Client).\footnote{\url{https://elasticsearch-py.readthedocs.io/en/master/}.}
We apply the default tokenizers and analyzers for Arabic, Bengali, Finnish and Russian. 
Japanese and Korean are not supported by the default ElasticSearch language analyzers, so we use Kuromoji\footnote{\url{https://www.elastic.co/guide/en/elasticsearch/plugins/7.9/analysis-kuromoji.html}.} and Nori plugins\footnote{\url{https://www.elastic.co/guide/en/elasticsearch/plugins/7.9/analysis-nori.html}.} for Japanese and Korean respectively.
Note that we do not implement a BM25-based retriever for Telugu, since it is not supported by the default language analyzer and we could not find an official plugin for Telugu.

\subsection{Machine Reading Models}
We use the official hyperparameters for machine reading components of DPR and Path Retriever. 
Table~\ref{tab:qa_hyp} shows the list of the hyperparameters used to train a multilingual machine reading model for the monolingual pipeline in \xorfull. We lowercased input paragraphs and questions. 
\begin{table}[t!]
\centering
\small
\begin{tabular}{@{} l@{\hspace{-0.0cm}} r @{}}
\toprule[.1em]
\textbf{Hyperparameter} & \textbf{Value}\\
\midrule[.1em]
max sequence length & 384\\
document stride & 128 \\
max query length & 64 \\
Adam lrate& $5\times 10^{-4}$ \\
Adam $\epsilon$&  $1\times 10^{-8}$\\
max gradient norm& 1.0 \\
\# train epochs & 3.0 \\
seed & 42 \\
\bottomrule[.1em]
\end{tabular}
 \caption{Hyperparameters for our machine reading model in the monolingual pipeline.}
 \label{tab:qa_hyp}
\end{table}

\paragraph{Choice of negative and positive examples.}
For the Path Retriever and BM25 baselines' reader, we sample three negative paragraphs per annotated question-gold paragraph pair and train a model that jointly predicts an answer span
and relevance score of each paragraph to the
question, following \citet{Asai2020Learning}. 
In DPR, the training examples are retrieved by the trained retriever, and we train the reader with 24 negative paragraphs by distant supervision \cite{karpukhin2020dense}. 
We use human translated English questions to train English reader models, and use the original questions in $L_i$ to train a multilingual reader model. 

\section{Additional Results and Analysis}
\label{ap_sec:additional_results}
\subsection{Additional Experimental Results}
\paragraph{\xorretrieve.}
We present the R@2kt scores of the retrieval baselines in Table~\ref{tab:main_results_retr_2k}. 
{As shown in Table~\ref{tab:main_results_retr}, given human translations, DPR generally outperforms other two retrieval baselines. }
\begin{table}[t!]
\small
\addtolength{\tabcolsep}{-4.0pt}

    \centering
    \begin{tabular}{l|ccc|cc|cc|c}
\toprule
 & \multicolumn{3}{c|}{Human} &  \multicolumn{2}{c|}{GMT}  &  \multicolumn{2}{c|}{Our MT} & \textit{Multi.} \\
& DPR & \textsc{Path} & BM &  DPR  & \textsc{Path} & DPR  & \textsc{Path} & DPR \ \\ \midrule
Ar & \textbf{65.8}  &  65.0 & 41.6 &  61.7 & 59.1 & 48.3 & 45.0 & 41.2 \\
Bn & 72.8 & \textbf{78.1}  & 57.7 & 72.0 & 58.2 & 54.4 & 60.9 & 43.9 \\
Fi  &  66.5&  \textbf{68.0} & 43.7 & 60.6  &  60.3 & 56.7 & 56.6 & 50.3 \\
Ja  & \textbf{62.0} & 59.0 & 38.8 & 52.1&  50.0&  41.8 & 36.7&  29.1 \\
Ko  &  \textbf{65.0} & 60.0 & 43.8 &  57.9 & 50.3 & 39.4& 33.8 & 34.5 \\
Ru & 57.5 & \textbf{59.9} &  35.2 & 51.2& 54.1 &  39.6& 34.7  & 35.3 \\
Te &  \textbf{66.3}& 59.6 & 44.6 & 59.4& 58.0 & 18.7 & 15.7 & 37.2 \\
\midrule
Av.\ &\textbf{65.1} & 64.3 &  43.5 & 59.3& 58.2& 42.7 & 40.5 & 38.8 \\
 \bottomrule
    \end{tabular}
    \caption{R@2kt (\S\ref{sec:sub_task_retrieval}) on the test data in the \xorretrieve setting. \textsc{Path} and BM denote Path Retriever and BM25 respectively.
    The rightmost column is a multilingual approach that bypasses the query translation step (\S\ref{sec:sub_task_retrieval}).
    }
    \label{tab:main_results_retr_2k}
\end{table}
We also present R@2kt and R@5kt of our DPR models on our development set in Table~\ref{tab:main_results_retr_dev}, and we observe a similar performance trend to the test set: models with queries translated by GMT outperform other models in all of the \tydixor languages. 
Comparing the two baselines that do not use external black-box APIs, we see that the translation approach (Our MT) outperforms the multilingual one (\textit{Multi}.) in Arabic, Bengali, Finnish Japanese, and Korean, while it performs poorly in Telugu.
These results are consistent with the ones on the test data in Table. \ref{tab:main_results_retr}.

\begin{table}[t!]
\small
\addtolength{\tabcolsep}{-4.0pt}
    \centering
    \begin{tabular}{l|cc|cc|cc}
\toprule
 &\multicolumn{2}{c|}{GMT}  &  \multicolumn{2}{c|}{Our MT} & \multicolumn{2}{c}{\textit{Multi.}} \\
& R@2kt &R@5kt & R@2kt &R@5kt & R@2kt &R@5kt \\ \midrule
Ar  & 62.5 & 69.6& 43.4 & 52.4 & 38.8 & 48.9 \\
Bn & 74.7& 82.2& 53.9& 62.8 & 48.4 & 60.2 \\
Fi & 57.3& 62.4& 55.1& 61.8 & 52.5 & 59.2 \\
Ja & 55.6& 64.7 &40.2 & 48.1 &  26.6 & 34.9 \\
Ko & 60.0 &  68.8 & 50.5 & 58.6 & 44.2 & 49.8 \\
Ru & 52.7& 60.8& 30.8 & 37.8 & 33.3 & 43.0 \\
Te & 72.3 & 79.0& 20.2 & 32.4 & 39.9 & 55.5 \\
\midrule
Av.\ & 62.2 & 69.6 & 42.0&  50.6 &   40.5 & 50.2 \\
 \bottomrule
    \end{tabular}
    \caption{R@5kt (\S\ref{sec:sub_task_retrieval}) of DPR models (translate DPR and multilingual DPR) on the development data in the \xorretrieve setting. 
    }
    \label{tab:main_results_retr_dev}
\end{table}

\paragraph{\xorengspan.}
Table~\ref{tab:main_results_eng_span_dev} shows the F1 and EM scores of our DPR models on the development data in the \xorengspan~setting. 
Similar to the results on \xorretrieve, GMT significantly outperforms our MT and our multilingual model. Probably due to the error propagation, the Telugu performance of our MT baseline is low, indicating the importance of developing a multilingual baseline that could perform well on languages with little parallel data for translation training. 
\begin{table}[t!]
\small
    \centering
    \begin{tabular}{l|cc|cc|cc}
\toprule
 &\multicolumn{2}{c|}{GMT}  &  \multicolumn{2}{c|}{Our MT} & \multicolumn{2}{c}{\textit{Multi.}} \\
& F1 & EM & F1 & EM & F1 & EM \\ \midrule
Ar  & 35.4 & 27.7 & 20.9 & 14.9 & 17.2 & 12.3 \\
Bn &  42.1& 35.3 & 25.2& 20.5  & 21.8 & 17.3  \\
Fi & 31.8& 23.1 & 31.9 & 23.3 & 27.6& 20.8  \\
Ja & 27.2& 20.9 & 19.6 & 15.5 & 15.5&  12.8 \\
Ko & 32.5 & 22.7 & 25.3 & 18.1 & 18.5& 14.4 \\
Ru & 34.7& 28.2 & 16.1 &  11.4& 21.3  & 17.3 \\
Te & 35.0 & 27.4 & 3.6 &  1.7  & 17.7 & 13.1 \\
\midrule
Av.\ & 35.0 & 27.4 & 20.4 &  15.1 &  19.9 & 15.4 \\
 \bottomrule
    \end{tabular}
    \caption{F1 and EM scores of our DPR models (translate DPR and multilingual DPR) on the development data in the \xorengspan setting. 
    }
    \label{tab:main_results_eng_span_dev}
\end{table}

\paragraph{\xorfull.}
We present F1, BLEU and EM scores for \xorfull in Tables~\ref{tab:main_results_full_f1_all}, \ref{tab:main_results_full_em} and \ref{tab:main_results_full_bleu}. 
We also present F1 scores and average F1, BLEU and EM scores on the development set in Table~\ref{tab:main_results_full_all_dev}.
\begin{table*}
\addtolength{\tabcolsep}{-1.8pt}
\small
    \centering
    \begin{tabular}{l|ll|ll|ccccccc|c}
\toprule
Wiki& \multicolumn{2}{c|}{Translation} &\multicolumn{2}{c|}{Retrieval} & \multicolumn{7}{c}{Target Language $L_i$} &   \\
Corpus & Query & Answer & $L_i$ & Eng.\ & {\bf Ar} & {\bf Bn} & {\bf Fi} & {\bf Ja} & {\bf Ko} & {\bf Ru} & {\bf Te} & {\bf Avg.} \\ \midrule
   \multirow{4}{*}{$\mathbf{W}_{i, {eng}}$} & GMT & GMT & GS  & DPR &  \textbf{31.5} & 19.0 & 18.3 & 8.8 & \textbf{20.1} & \textbf{19.8} & \textbf{13.6} & 18.7  \\
& Our MT & Our MT & GS & DPR & 29.6 & 6.6 & 15.5 & 7.6 & 16.4 & 18.7 & 1.7 & 13.7  \\
& Our MT & Our MT &  BM25 & DPR & 12.1 & \textbf{22.0} & 9.3 & 5.4 & 9.7 & 7.4 & 0.8 & 9.5  \\
& -- & GMT & GS & mDPR & 30.5 & 5.2 & 16.9 & 8.2& 17.6 & \textbf{19.8} & 6.0 & 15.7 \\
 \midrule
\multirow{2}{*}{$W_{eng}$} & GMT & GMT & -- & DPR  & 23.9 & 18.5 & \textbf{22.9}  & \textbf{24.1} & 17.5 & 16.8& 13.2 & \textbf{19.5} \\
& Our MT& Our MT & -- & DPR  & 7.6 & 5.9  & 16.2  & 9.0 & 5.3 & 5.5 & 0.8 & 7.2 \\
& -- & GMT & -- & mDPR  & 12.4 & 9.7  & 19.1  & 14.0 & 8.2 & 10.9& 5.4 & 11.3 \\
\hdashline
\multirow{2}{*}{$W_{i}$}  & -- & -- & GS & --  & 29.0 & 0.9 & 9.5 & 6.2 & 14.3& 18.5 & 0.9 & 11.3 \\
& -- & -- &  BM25& -- & 12.0 & \textbf{22.0} & 9.3 & 5.3 & 9.7 & 7.4 & -- & -- \\
 \bottomrule
    \end{tabular}
    \caption{Performance on \xorfull task (F1 scores on the test data). 
    ``GS'' denotes Google Search retrieval.
    The bottom section shows results from single Wikipedia baselines.
    ElasticSearch for BM25 does not support Telugu. 
    ``mDPR'' denotes a DPR model where query and context encoders are initialized with multilingual BERT. 
    }
    \label{tab:main_results_full_f1_all}
\end{table*}
\begin{table*}
\addtolength{\tabcolsep}{-1.8pt}
\small
    \centering
    \begin{tabular}{l|ll|ll|ccccccc|c}
\toprule
Wiki& \multicolumn{2}{c|}{Translation} &\multicolumn{2}{c|}{Retrieval} & \multicolumn{7}{c}{Target Language $L_i$} \\
Corpus & Query & Answer & $L_i$ & Eng.\ & {\bf Ar} & {\bf Bn} & {\bf Fi} & {\bf Ja} & {\bf Ko} & {\bf Ru} & {\bf Te}  &{\bf Avg.} \\ \midrule
   \multirow{4}{*}{$\mathbf{W}_{i, {eng}}$} & GMT & GMT & GS  & DPR &  \textbf{22.1} & 10.9 & 13.3 & 3.0 & \textbf{20.1} & \textbf{11.4} & \textbf{9.1} &  \textbf{12.1} \\
& Our MT & Our MT & GS & DPR & 20.9 & 2.2 & 10.9 & 2.3 & 12.6 & 10.5 & 1.4 & 8.7 \\
& Our MT  & Our MT &  BM25 & DPR & 7.7 & \textbf{15.4} & 6.4 & 1.3 & 6.7 & 3.9 &  0.6  & 6.0  \\
& -- & GMT & GS & mDPR  & 21.4& 5.2 & 12.1  & 2.7 & 13.3 & 11.3 & 3.9 & 10.0  \\
 \midrule
\multirow{2}{*}{$W_{eng}$} & GMT & GMT & -- & DPR  & 12.3 & 10.1 & \textbf{16.6}  & \textbf{14.1} & 11.5 & 10.4 & 8.5 & 12.0 \\
& Our MT& Our MT & -- & DPR  & 2.5 & 1.5 & 10.3  & 3.3 & 2.9 &  2.5 & 0.5 & 3.4  \\
& -- & GMT & -- & mDPR  & 6.7 & 4.5 & 13.5  & 8.1 & 8.2 & 6.5& 3.1 & 6.8 \\
\hdashline
\multirow{2}{*}{$W_{i}$}  & -- & -- & GS & --  & 20.6 & 0.7 & 7.1 & 1.5 & 11.5 & 10.4 & 0.8 & 7.5 \\
& -- & -- &  BM25& -- & 7.7  & 15.3  &6.4 & 1.3 & 6.7& 3.9 & -- & --  \\
 \bottomrule
    \end{tabular}
    \caption{Performance on \xorfull (EM scores on the test data). ``GS'' denotes Google Search retrieval. The bottom section shows results from single Wikipedia baselines. ElasticSearch for BM25 does not support Telugu. ``mDPR'' denotes a DPR model where query and context encoders are initialized with multilingual BERT.}
    \label{tab:main_results_full_em}
\end{table*}

\begin{table*}
\addtolength{\tabcolsep}{-1.8pt}
\small
    \centering
    \begin{tabular}{l|ll|ll|ccccccc|c}
\toprule
Wiki& \multicolumn{2}{c|}{Translation} &\multicolumn{2}{c|}{Retrieval} & \multicolumn{7}{c}{Target Language $L_i$} \\
Corpus & Query & Answer & $L_i$ & Eng.\ & {\bf Ar} & {\bf Bn} & {\bf Fi} & {\bf Ja} & {\bf Ko} & {\bf Ru} & {\bf Te} &  {\bf Avg.} \\ \midrule
   \multirow{4}{*}{$\mathbf{W}_{i, {eng}}$} & GMT & GMT & GS  & DPR & \textbf{29.7} & 22.1 & 18.8 & 2.2 & \textbf{13.3} & \textbf{18.0} & \textbf{13.5} & \textbf{16.8} \\
& Our MT & Our MT & GS & DPR & 27.8 &  7.4 & 10.9 & 2.0 & 12.6 & 17.0 & 1.1 & 8.9 \\
& Our MT & Our MT &  BM25 & DPR &  12.8 & \textbf{22.9}  & 6.4 & 1.2 & 7.0 & 7.3  & 0.3 & 12.0 \\
& -- & GMT & GS & mDPR & 27.8 & 7.0  &  13.9 & 1.8 & 11.3 & 17.0 & 5.3 & 13.9  \\
 \midrule
\multirow{2}{*}{$W_{eng}$} & GMT & GMT & -- & DPR  & 24.5 & 21.4 & \textbf{20.6} & \textbf{6.1} & 10.0 & 14.2 & 13.3 & 15.7 \\
& Our MT& Our MT & -- & DPR  & 8.6 &  6.7&  16.7  & 2.8 & 3.9 & 5.0 & 0.3 & 6.3 \\
& -- & GMT & -- & mDPR  & 12.2 & 10.2 & 16.7  & 2.4 & 4.7 & 8.2 & 8.3 & 9.0 \\
\hdashline
\multirow{2}{*}{$W_{i}$}  & -- & -- & GS & --  & 27.3 & 0.7 & 10.4 & 1.6 & 10.4 & 16.6 & 0.8  & 9.7 \\
& -- & -- &  BM25& -- & 12.8 &\textbf{22.9} & 10.6 & 1.2 & 7.0 & 7.3 & -- & -- \\
 \bottomrule
    \end{tabular}
    \caption{Performance on \xorfull (BLEU scores on the test data). 
    ``GS'' denotes Google Search retrieval.
    The bottom section shows results from single Wikipedia baselines.
    ElasticSearch for BM25 does not support Telugu.
    ``mDPR'' denotes a DPR model where query and context encoders are initialized with multilingual BERT. 
    }
    \label{tab:main_results_full_bleu}
\end{table*}
\begin{table*}
\addtolength{\tabcolsep}{-1.8pt}
\small
    \centering
    \begin{tabular}{ll|ll|ccccccc|ccc}
\toprule
 \multicolumn{2}{c|}{Translation} &\multicolumn{2}{c|}{Retrieval} & \multicolumn{7}{c}{Target Language $L_i$} & \multicolumn{3}{|c}{Macro Average}\\
 Query & Answer & $L_i$ & Eng.\ & {\bf Ar} & {\bf Bn} & {\bf Fi} & {\bf Ja} & {\bf Ko} & {\bf Ru} & {\bf Te} & {\bf F1} & {\bf EM} & {\bf BLEU} \\ \midrule
GMT & GMT & GS  & DPR & \textbf{18.0} & \textbf{29.1} & 13.8 & \textbf{5.7} & \textbf{15.2} & 14.9 & \textbf{15.6} & \textbf{16.0} & \textbf{9.9} & \textbf{14.9} \\
Our MT & Our MT & GS & DPR & 17.7 &  4.5 & 13.0 & \textbf{5.7}& 15.0&  14.9 & 8.8  & 11.4 & 6.3 & 10.3 \\
Our MT & Our MT &  BM25 & DPR & 9.2 & 15.8 & \textbf{14.4} & 4.8 & 7.9 & 5.2 & 0.5 & 8.3 & 4.6 & 7.5  \\
-- & GMT & GS & mDPR & 17.8 & 15.3 & 12.6 & 5.6& \textbf{15.2} & \textbf{15.0} & 10.1  &13.1 & 7.7 & 12.2  \\
 \bottomrule
    \end{tabular}
    \caption{Performance on \xorfull (dev data F1 scores and average F1, EM and BLEU scores). 
    ``GS'' denotes Google Search retrieval, and ``mDPR'' denotes a DPR model where query and context encoders are initialized with multilingual BERT.}
    \label{tab:main_results_full_all_dev}
\end{table*}

\subsection{Additional Analysis}
\label{ap_dec:additional_analysis}
\paragraph{Single language Wikipedia ablations in \xorfull.}
In \xorfull, a system is expected to answer a question in the target language by consulting multilingual Wikipedia corpora, but which language answer content exists in is not known \textit{a priori} (\S\ref{sec:main_task_definition}). 
To understand the benefit of retrieving evidence from a multilingual document pool, we run single language Wikipedia ablations. In this study, we conduct ablations in which systems only use either English Wikipedia ($W_{eng}$) or the target language's Wikipedia ($W_i$).
We run the monolingual baselines for $W_{i}$ only and the cross-lingual baselines for $W_{eng}$ only. For the cross-lingual baseline, all predicted answers will be translated back to the target languages.

The bottom section of Table \ref{tab:main_results_full_f1_all} shows the full results of single language Wikipedia ablations on \xorfull.
In a majority of the languages, we observed performance drops from the full models that use both $W_{eng}$ and $W_i$ (e.g., 20.1 vs.\ 14.3 F1 in Korean). In Japanese and Finnish, our English Wikipedia only baselines outperform the full models. Currently, the answer aggregation process prioritizes answers predicted by monolingual models, but the monolingual models perform poorly in those two languages. Future work can address the challenges of improving evidence and answer aggregation from multilingual document collections.

\paragraph{Per-difficulty retrieval performance.}
\begin{table}[ht!]
\addtolength{\tabcolsep}{-0.8pt}

\small
    \centering
    \begin{tabular}{l|cc|cc}
\toprule
Query &  \multicolumn{2}{c}{Easy} & \multicolumn{2}{c}{Hard}\\
Translator & R@2kt & R@5kt & R@2kt &  R@5kt  \\ \midrule
{Human} & 65.3  & 72.5  & 59.9 & 68.9 \\
{GMT}&  61.1 &  67.7 &  54.3&  63.4  \\
{Our MT} & 41.9 &  49.9 &  37.7 &  44.5  \\\hdashline
{\it Multilingual} &  34.3 & 44.3  &  36.1 & 40.9  \\
 \bottomrule
    \end{tabular}
    \caption{Macro-averaged retrieval recall on the {\it easy} and {\it hard} subsets of the development set. 
    All configurations use DPR for retrieval.
    The \textit{Multilingual} model avoids query translation.
    }
    \label{tab:per_dficciult_retrieval}
\end{table}

We split our data by annotation difficulty i.e., whether or not a gold paragraph is selected by the BERT retriever used during annotation in our our collaborative annotation framework (\S\ref{sec:qa_annot_dataset}).
Table~\ref{tab:per_dficciult_retrieval} presents retrieval performance broken down by difficulty. 
We observed a large performance gap between the easy and hard subsets (65.3 for easy vs.\ 59.9 for hard subsets in R@2kt with human translation and DPR), suggesting that the questions from the hard subset are clearly more challenging than the ones from the easy subset.

\subsection{Qualitative Analysis on Translation Errors}
\label{ap_sec:trans_errors}
One primary challenge in question translation is precisely translating key words (e.g., entities, year); our MT correctly translates a Japanese question, \begin{CJK}{UTF8}{min}アーモンドアイはいつ生まれた？(\textit{When was Almond Eye born}; Almond Eye is a Japanese popular race horse\end{CJK})\footnote{\url{https://en.wikipedia.org/wiki/Almond_Eye}.} while Helsinki \cite{TiedemannThottingal:EAMT2020} translates it to ``When was almond born?'' This resulted in retrieval errors, and Wikipedia articles related to almonds were selected. Intrinsic metrics such as BLEU would not consider the importance of these translation mistakes.

\end{document}